\documentclass[runningheads]{llncs}
\usepackage{graphicx}
\usepackage{comment}
\usepackage{amsmath,amssymb,amsfonts,amssymb} %
\usepackage{color}
\usepackage{booktabs}
\usepackage{xspace}
\usepackage{enumitem}
\usepackage{microtype}
\usepackage{subcaption}
\usepackage{cite}
\usepackage{soul}
\usepackage[pagebackref=true,breaklinks=true,letterpaper=true,colorlinks,bookmarks=false]{hyperref}

\newcommand{\bp}{\mathbf{p}}

\newcommand{\bx}{\mathbf{x}}

\newcommand{\nR}{\mathbb{R}}

\newcommand{\cL}{\mathcal{L}}

\newcommand{\figref}[1]{Fig.~\ref{#1}}
\newcommand{\secref}[1]{Section~\ref{#1}}

\newcommand{\tabref}[1]{Table~\ref{#1}}

\makeatletter
\DeclareRobustCommand\onedot{\futurelet\@let@token\@onedot}
\def\@onedot{\ifx\@let@token.\else.\null\fi\xspace}
\def\eg{e.g\onedot} 
\def\ie{i.e\onedot}

\def\etal{et~al\onedot}

\makeatother

\newcommand{\boldparagraph}[1]{\vspace{0.2cm}\noindent{\bf #1:} }

\definecolor{darkgreen}{rgb}{0,0.7,0}
\definecolor{lightred}{rgb}{1.,0.5,0.5}

\usepackage[width=122mm,left=12mm,paperwidth=146mm,height=193mm,top=12mm,paperheight=217mm]{geometry}

\begin{document}
\pagestyle{headings}
\mainmatter
\def\ECCVSubNumber{597}  %

\title{Convolutional Occupancy Networks} %

\titlerunning{Convolutional Occupancy Networks}
\author{Songyou Peng\inst{1, 2} \quad Michael Niemeyer\inst{2,3} \quad Lars Mescheder\inst{2, 4}\thanks{This work was done prior to joining Amazon.}\\ Marc Pollefeys\inst{1, 5} \quad Andreas Geiger\inst{2,3}\\
}
\institute{$^1$ETH Zurich \quad $^2$Max Planck Institute for Intelligent Systems, T\"{u}bingen\\
$^3$University of T\"{u}bingen \quad $^4$Amazon, T\"{u}bingen \quad $^5$Microsoft}
\authorrunning{S. Peng et al.}

\maketitle

\begin{abstract}
Recently, implicit neural representations
have gained popularity for learning-based 3D reconstruction.
While demonstrating promising results, most implicit approaches are limited to comparably simple geometry of single objects and do not scale to more complicated or large-scale scenes. 
The key limiting factor of implicit methods is their simple fully-connected network architecture
which does not allow for integrating local information in the observations or incorporating inductive biases such as translational equivariance.
In this paper, we propose Convolutional Occupancy Networks, a more flexible implicit representation for detailed reconstruction of objects and 3D scenes.
By combining convolutional encoders with implicit occupancy decoders, our model incorporates inductive biases, enabling structured reasoning in 3D space.
We investigate the effectiveness of the proposed representation by reconstructing complex geometry from noisy point clouds and low-resolution voxel representations.
We empirically find that our method enables the fine-grained implicit 3D reconstruction of single objects, scales to large indoor scenes, and generalizes well from synthetic to real data.
\end{abstract}

\section{Introduction}\label{sec:intro}

\begin{figure}[!t]
	\begin{minipage}{0.41\linewidth}
		\begin{subfigure}{\linewidth}
			\hspace{0.03cm}\includegraphics[width=0.92\linewidth]{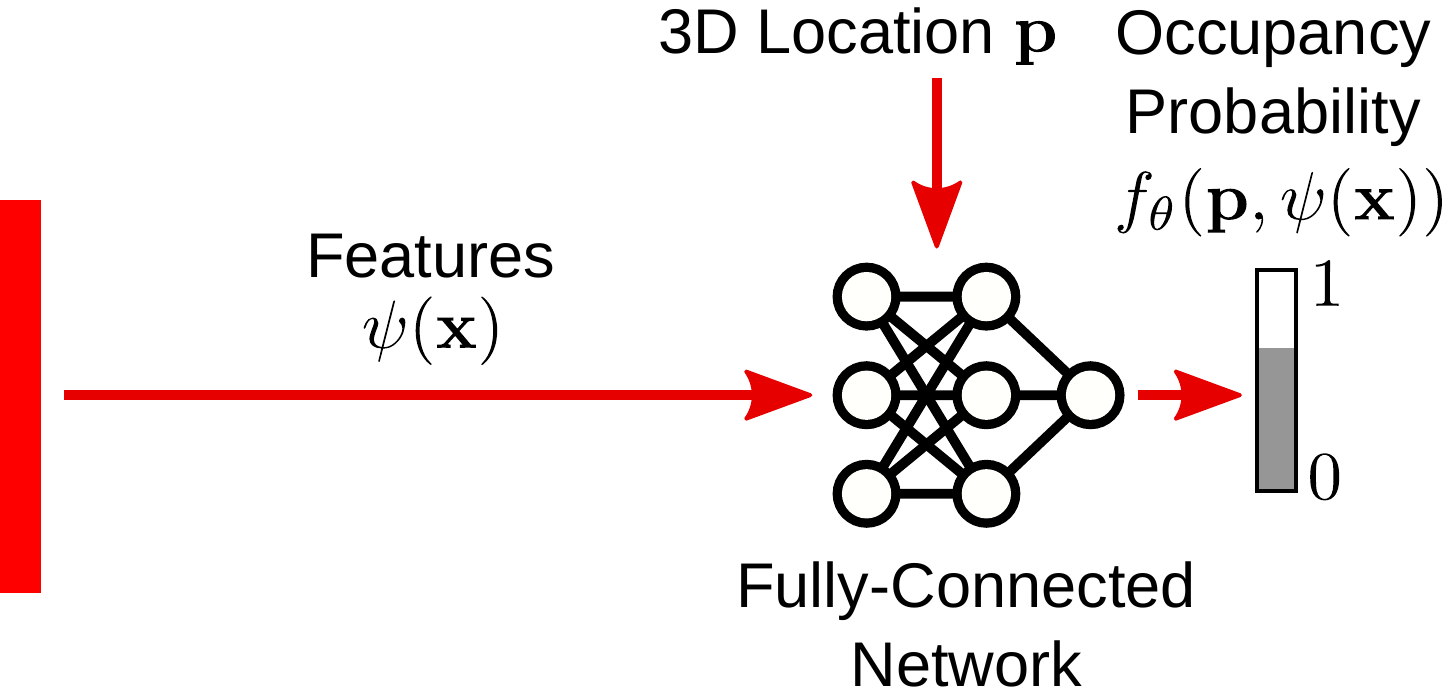}
			\caption{Occupancy Network \cite{Mescheder2019CVPR}}
			\vspace{0.4cm}
			\label{fig:teaser_a}
		\end{subfigure}
		\begin{subfigure}{\linewidth}
			\includegraphics[width=0.95\linewidth]{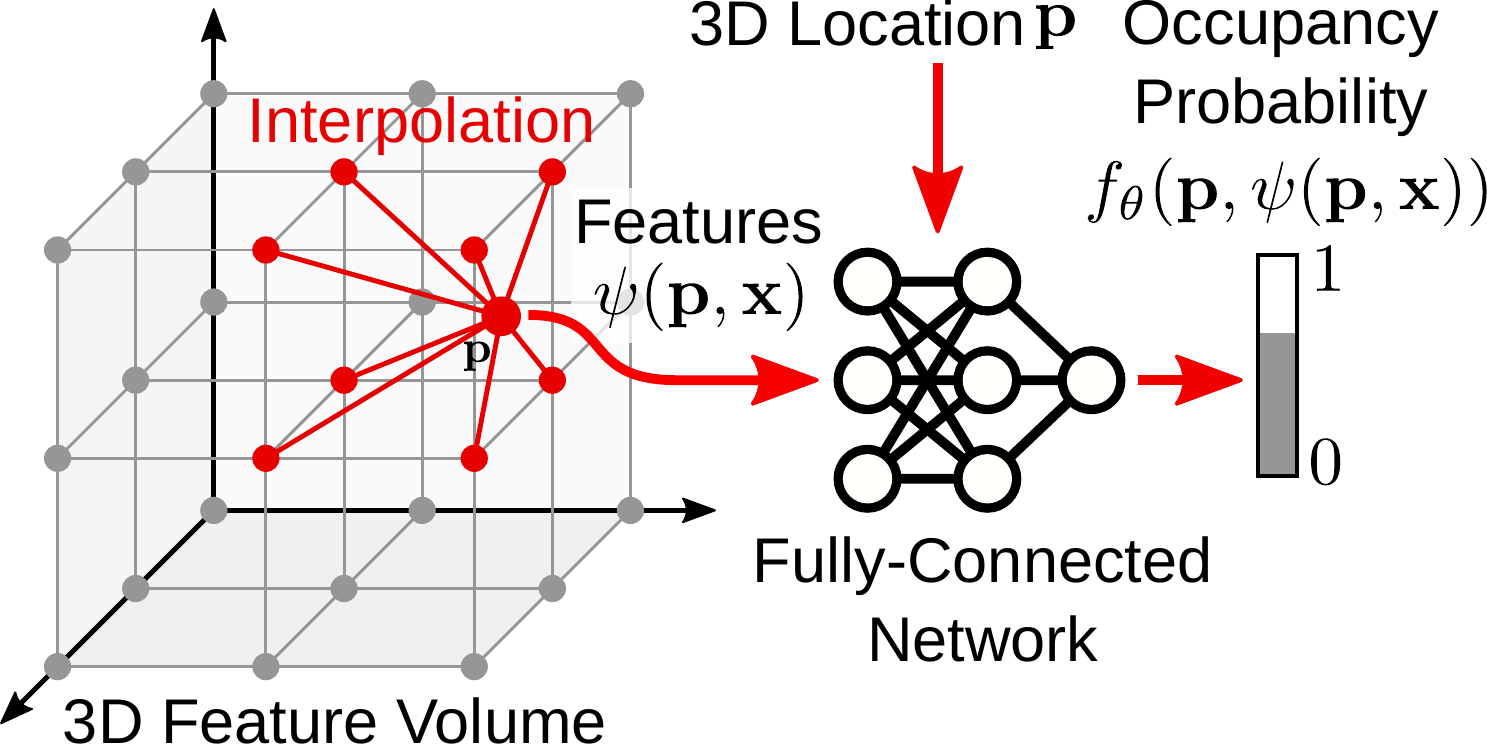}
			\vspace{-0.05cm}
			\caption{Conv. Occupancy Network}
			\label{fig:teaser_b}
		\end{subfigure}
	\end{minipage}\hfill%
	\begin{minipage}{0.58\linewidth}
		\begin{subfigure}{\linewidth}
			\includegraphics[width =1\linewidth]{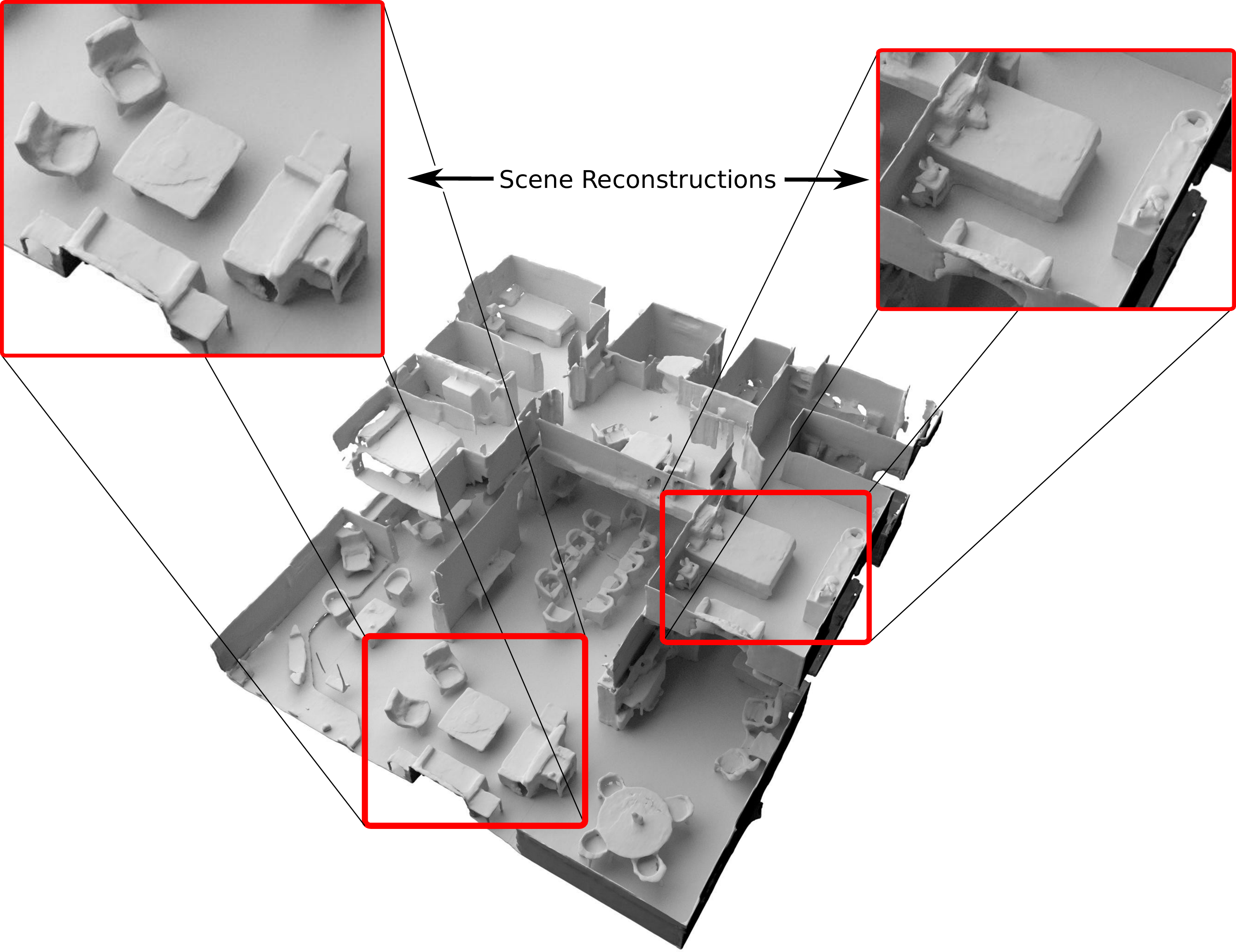}
			\caption{Reconstruction on Matterport3D~\cite{Chang2017THREEDV}}
			\label{fig:teaser_c}
		\end{subfigure}		
	\end{minipage}
\caption{\textbf{Convolutional Occupancy Networks.}
Traditional implicit models (\subref{fig:teaser_a}) are limited in their expressiveness due to their fully-connected network structure. We propose Convolutional Occupancy Networks (\subref{fig:teaser_b}) which exploit convolutions, resulting in scalable and equivariant implicit representations.
We query the convolutional features at 3D locations $\bp\in\nR^3$ using linear interpolation. In contrast to Occupancy Networks (ONet)~\cite{Mescheder2019CVPR}, the proposed feature representation $\psi(\bp,\bx)$ therefore depends on \textit{both} the input $\bx$ and the 3D location $\bp$. Fig.~(\subref{fig:teaser_c}) shows a reconstruction of a two-floor building from a noisy point cloud on the Matterport3D dataset~\cite{Chang2017THREEDV}.}
\label{fig:teaser}
\end{figure}

3D reconstruction is a fundamental problem in computer vision with numerous applications.
An ideal representation of 3D geometry should have the following properties:
a) encode complex geometries and arbitrary topologies,
b) scale to large scenes,
c) encapsulate local and global information, and
d) be tractable in terms of memory and computation.

Unfortunately, current representations for 3D reconstruction do not satisfy all of these requirements.
Volumetric representations \cite{Maturana2015IROS} are limited in terms of resolution due to their large memory requirements.
Point clouds \cite{Fan2017CVPR} are lightweight 3D representations but discard topological relations.
Mesh-based representations \cite{Groueix2018CVPR} are often hard to predict using neural networks.

Recently, several works~\cite{Mescheder2019CVPR,Park2019CVPR,Chen2019CVPR,Michalkiewicz2019ICCV} have introduced deep implicit representations which represent 3D structures using learned occupancy or signed distance functions.
In contrast to explicit representations, implicit methods do not discretize 3D space during training, thus resulting in continuous representations of 3D geometry without topology restrictions.
While inspiring many follow-up works~\cite{Sitzmann2019NIPS,Niemeyer2019ICCV,Oechsle2019ICCV,Genova2019ICCV,Genova2019ARXIV,Liu2019ARXIV,Niemeyer2019ARXIV,Liu2019NIPSb}, all existing approaches are limited to single objects and do not scale to larger scenes.
The key limiting factor of most implicit models is their simple fully-connected network architecture \cite{Mescheder2019CVPR,Park2019CVPR}
which neither allows for integrating local information in the observations, nor for incorporating inductive biases such as translation equivariance into the model.
This prevents these methods from performing \textit{structured reasoning} as they only act globally and result in overly smooth surface reconstructions.

In contrast, translation equivariant convolutional neural networks (CNNs) have demonstrated great success across many 2D recognition tasks including object detection and image segmentation.
Moreover, CNNs naturally encode information in a hierarchical manner in different network layers~\cite{Zeiler2014ECCV,Zhang2018CVPR}.
Exploiting these inductive biases is expected to not only benefit 2D but also 3D tasks, \eg, reconstructing 3D shapes of multiple similar chairs located in the same room.
In this work, we seek to combine the complementary strengths of convolutional neural networks with those of implicit representations.

Towards this goal, we introduce \emph{Convolutional Occupancy Networks}, a novel representation for accurate large-scale 3D reconstruction\footnote{With 3D reconstruction, we refer to 3D surface reconstruction throughout the paper.} with continuous implicit representations (\figref{fig:teaser}). 
We demonstrate that this representation not only preserves fine geometric details, but also enables the reconstruction of complex indoor scenes at scale.
Our key idea is to establish rich input features, incorporating inductive biases and integrating local as well as global information.
More specifically, we exploit convolutional operations to obtain translation equivariance and exploit the local self-similarity of 3D structures.
We systematically investigate  multiple design choices, ranging from canonical planes to volumetric representations.
Our contributions are summarized as follows:
\begin{itemize}
  \item We identify major limitations of current implicit 3D reconstruction methods.
  \item We propose a flexible translation equivariant architecture which enables accurate 3D reconstruction from object to scene level.
  \item We demonstrate that our model enables generalization from synthetic to real scenes as well as to novel object categories and scenes.
\end{itemize}
Our code and data are provided at \url{https://github.com/autonomousvision/convolutional_occupancy_networks}.

\section{Related Work}\label{sec:related-work}

Learning-based 3D reconstruction methods can be broadly categorized by the output representation they use.

\boldparagraph{Voxels}
Voxel representations are amongst the earliest representations for learning-based 3D reconstruction~\cite{Wu2015CVPR,Choy2016ECCV,Wu2016NIPS}.
Due to the cubic memory requirements of voxel-based representations, several works proposed to operate on multiple scales or use octrees for efficient space partitioning~\cite{Hane2017THREEDV,Tatarchenko2017ICCV,Riegler2017CVPR,Riegler2017THREEDV,Maturana2015IROS,Dai2017CVPRa}.
However, even when using adaptive data structures, voxel-based techniques are still limited in terms of memory and computation.

\boldparagraph{Point Clouds}
An alternative output representation for 3D reconstruction is 3D point clouds which have been used in~\cite{Fan2017CVPR,Lin2018AAAI,Yang2019ICCV,Prokudin2019ICCV}.
However, point cloud-based representations are typically limited in terms of the number of points they can handle. Furthermore, they cannot represent topological relations.

\boldparagraph{Meshes}
A popular alternative is to directly regress the vertices and faces of a mesh \cite{Gkioxari2019ICCV,Groueix2018CVPR,Kanazawa2018ECCV,Wang2018ECCV,Wen2019ICCV,Lin2019CVPR,Liao2018CVPR} using a neural network.
While some of these works require deforming a template mesh of fixed topology, others result in non-watertight reconstructions with self-intersecting mesh faces.

\boldparagraph{Implicit Representations}
More recent implicit occupancy~\cite{Mescheder2019CVPR,Chen2019CVPR} and distance field~\cite{Park2019CVPR,Michalkiewicz2019ICCV} models use a neural network to infer an occupancy probability or distance value given any 3D point as input.
In contrast to the aforementioned explicit representations which require discretization (\eg, in terms of the number of voxels, points or vertices), implicit models represent shapes continuously and naturally handle complicated shape topologies.
Implicit models have been adopted for 
learning implicit representations from
 images~\cite{Liu2019NIPSb,Sitzmann2019NIPS,Liu2019ARXIV,Niemeyer2019ARXIV}, for encoding texture information~\cite{Oechsle2019ICCV}, for 4D reconstruction~\cite{Niemeyer2019ICCV} as well as for primitive-based reconstruction~\cite{Genova2019ICCV,Genova2019ARXIV,Jeruzalski2019ARXIV,Paschalidou2020CVPR}.
Unfortunately, all these methods are limited to comparably simple 3D geometry of single objects and do not scale to more complicated or large-scale scenes.
The key limiting factor is the simple fully-connected network architecture which does not allow for integrating local features or incorporating inductive biases such as translation equivariance.

Notable exceptions are PIFu~\cite{Saito2019ICCV} and DISN~\cite{Xu2019NIPS} which use pixel-aligned implicit representations to reconstruct people in clothing \cite{Saito2019ICCV} or ShapeNet objects~\cite{Xu2019NIPS}.
While these methods also exploit convolutions, all operations are performed in the \textit{2D image domain}, restricting these models to image-based inputs and reconstruction of single objects.
In contrast, in this work, we propose to aggregate features in \textit{physical 3D space}, exploiting both 2D and 3D convolutions.
Thus, our world-centric representation is independent of the camera viewpoint and input representation. Moreover, we demonstrate the feasibility of implicit 3D reconstruction at scene-level as illustrated in \figref{fig:teaser_c}.

In concurrent work, Chibane \etal \cite{Chibane2020CVPR} present a model similar to our convolutional volume decoder. In contrast to us, they only consider a single variant of convolutional feature embeddings (3D), use lossy discretization for the 3D point cloud encoding and only demonstrate results on single objects and humans, as opposed to full scenes. 
In another concurrent work, Jiang \etal~\cite{Jiang2020CVPR} leverage 
shape priors for scene-level implicit 3D reconstruction. In contrast to us, they use 3D point normals as input and require optimization at inference time.

\section{Method}\label{sec:method}

\begin{figure}[t]
	\begin{minipage}{0.4\linewidth}
		\begin{subfigure}{\linewidth}
			\includegraphics[width=\linewidth]{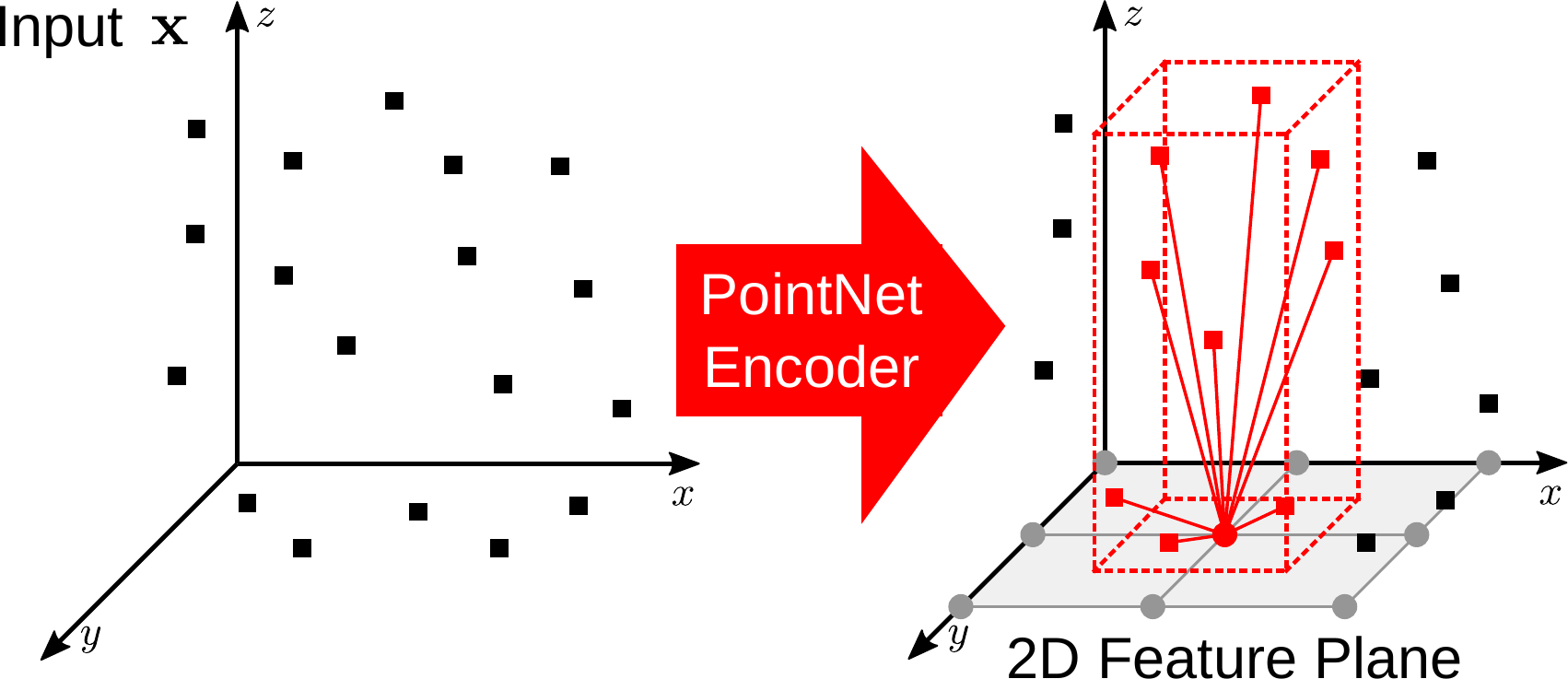}
			\vspace{-0.52cm}
			\caption{Plane Encoder}
			\vspace{0.8cm}
			\label{fig:model_a}
		\end{subfigure}
		\begin{subfigure}{\linewidth}
			\includegraphics[width=\linewidth]{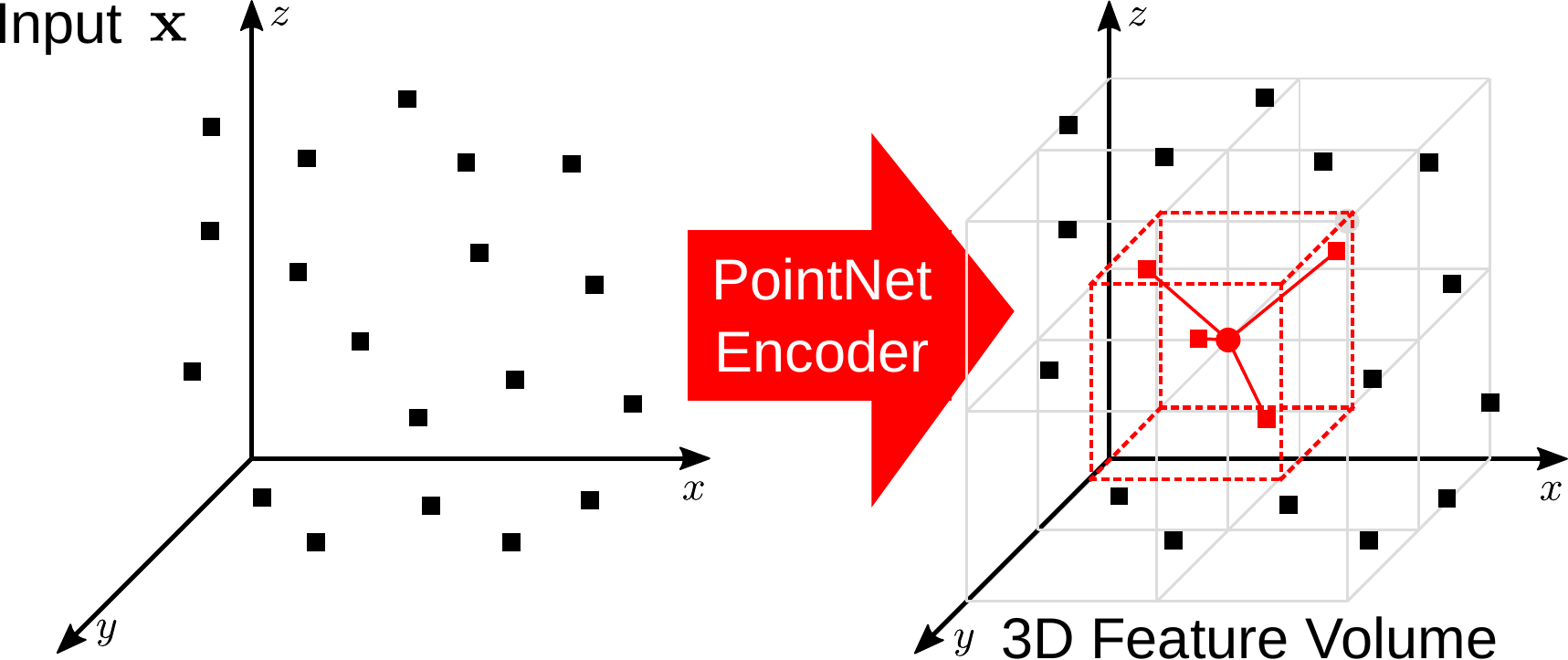}
			\vspace{-0.52cm}
			\caption{Volume Encoder}
			\label{fig:model_b}
		\end{subfigure}
	\end{minipage}
	~~~
	\begin{minipage}{0.01\linewidth}
	\includegraphics[height=8cm]{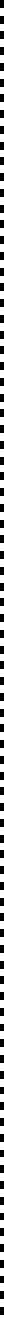}
	\end{minipage}
	~~
	\begin{minipage}{.52\linewidth}
		\begin{subfigure}{\linewidth}
			\includegraphics[width=\linewidth]{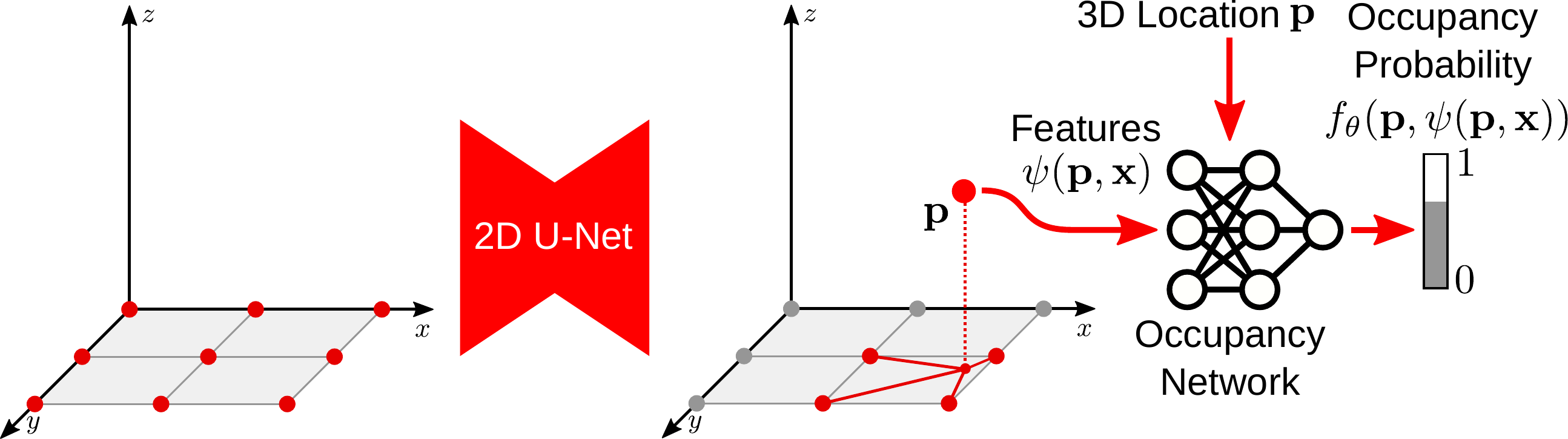}
			\vspace{-0.52cm}
			\caption{Convolutional Single-Plane Decoder}
			\vspace{0.4cm}
			\label{fig:model_c}
		\end{subfigure}
		\begin{subfigure}{\linewidth}
			\includegraphics[width=\linewidth]{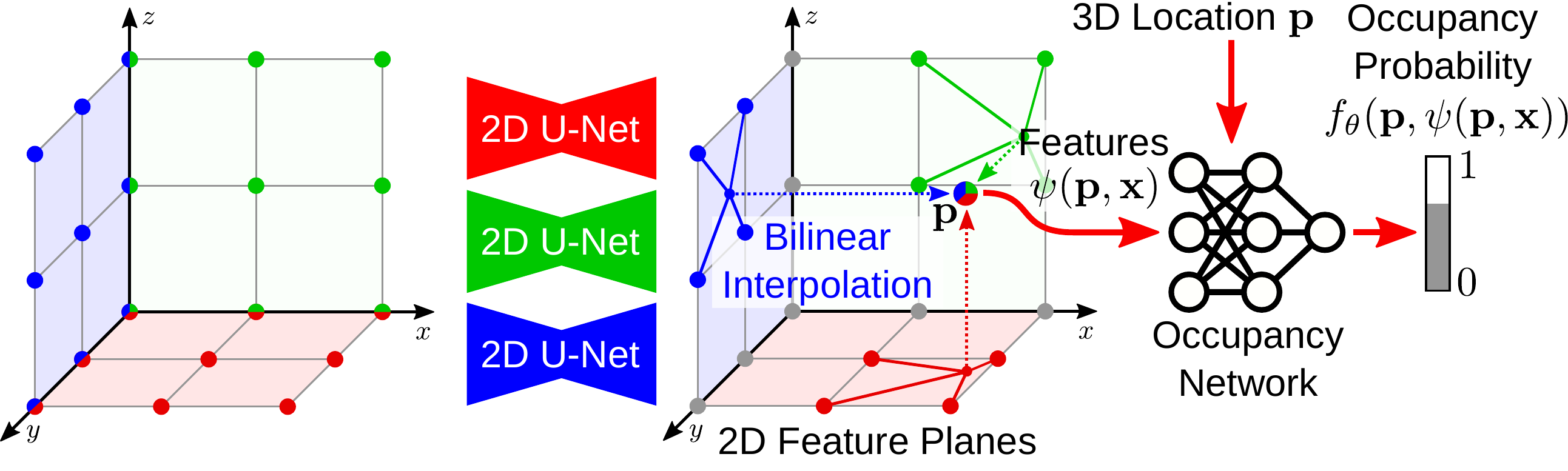}
			\vspace{-0.52cm}
			\caption{Convolutional Multi-Plane Decoder}
			\vspace{0.4cm}
			\label{fig:model_d}
		\end{subfigure}
		\begin{subfigure}{\linewidth}
			\includegraphics[width=\linewidth]{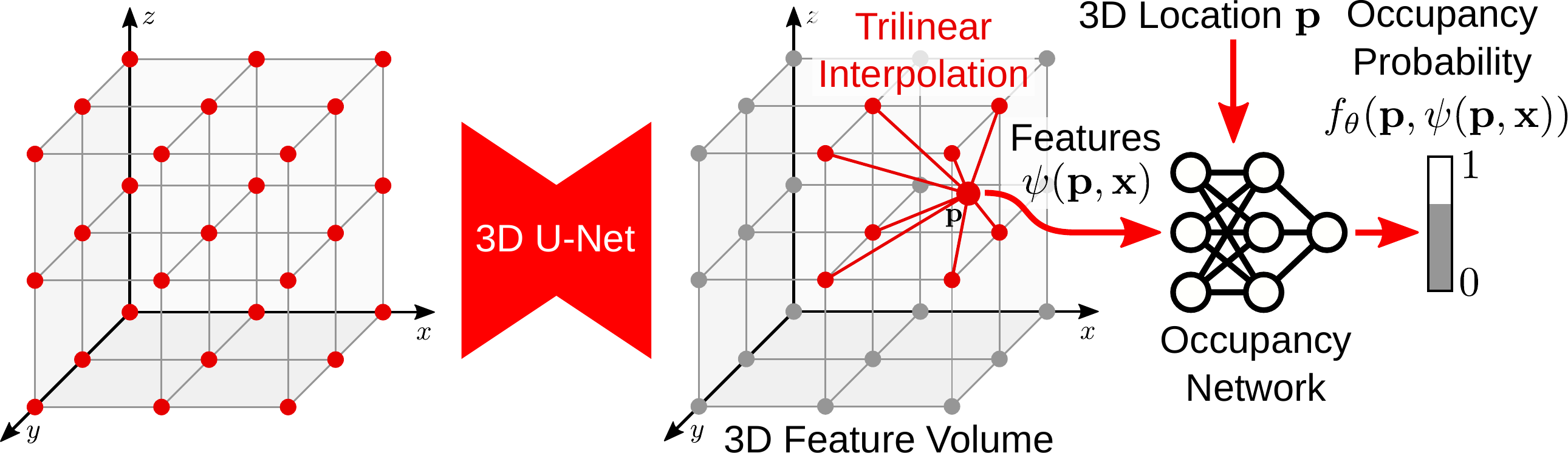}
			\vspace{-0.52cm}
			\caption{Convolutional Volume Decoder}
			\label{fig:model_e}
		\end{subfigure}
	\end{minipage}
	\caption{\textbf{Model Overview.}
	The \textbf{encoder (left)} first converts the 3D input $\bx$ (\eg, noisy point clouds or coarse voxel grids) into features using task-specific neural networks. Next, the features are projected onto one or multiple planes (\figref{fig:model_a}) or into a volume (\figref{fig:model_b}) using average pooling.
	The \textbf{convolutional decoder (right)} processes the resulting feature planes/volume using 2D/3D U-Nets to aggregate local and global information. For a query point $\bp \in \mathbb{R}^3$, the point-wise feature vector $\psi(\bx, \bp)$ is obtained via bilinear (\figref{fig:model_c} and \figref{fig:model_d}) or trilinear (\figref{fig:model_e}) interpolation. Given feature vector $\psi(\bx, \bp)$ at location $\bp$, the occupancy probability is predicted using a fully-connected network $f_\theta(\bp, \psi(\bp, \bx))$.}
	\vspace{-1em}
	\label{fig:model}
\end{figure}

Our goal is to make implicit 3D representations more expressive.
An overview of our model is provided in \figref{fig:model}.
We first encode the input $\bx$ (\eg, a point cloud) into a 2D or 3D feature grid (left).
These features are processed using convolutional networks and decoded into occupancy probabilities via a fully-connected network.
We investigate planar representations (\subref{fig:model_a}+\subref{fig:model_c}+\subref{fig:model_d}), volumetric representations (\subref{fig:model_b}+\subref{fig:model_e}) as well as combinations thereof in our experiments.
In the following, we explain the encoder (\secref{subsec:encoder}), the decoder (\secref{subsec:decoder}), the occupancy prediction (\secref{subsec:occupancy-prediction}) and the training procedure (\secref{subsec:training}) in more detail.

\subsection{Encoder}\label{subsec:encoder}

While our method is independent of the input representation, we focus on 3D inputs to demonstrate the ability of our model in recovering fine details and scaling to large scenes. More specifically, we assume a noisy sparse point cloud (\eg, from structure-from-motion or laser scans) or a coarse occupancy grid as input $\bx$.

We first process the input $\bx$ with a task-specific neural network to obtain a feature encoding for every point or voxel.
We use a one-layer 3D CNN for voxelized inputs, and a shallow PointNet~\cite{Qi2017CVPR} with local pooling for 3D point clouds.
Given these features, we construct planar and volumetric feature representations in order to encapsulate local neighborhood information as follows.

\boldparagraph{Plane Encoder}
As illustrated in \figref{fig:model_a}, for each input point, we perform an orthographic projection onto a canonical plane (\ie, a plane aligned with the axes of the coordinate frame) which we discretize at a resolution of $H \times W$ pixel cells.
For voxel inputs, we treat the voxel center as a point and project it to the plane.
We aggregate features projecting onto the same pixel using average pooling, resulting in planar features with dimensionality $H \times W \times d$, where $d$ is the feature dimension.

In our experiments, we analyze two variants of our model: one variant where features are projected onto the ground plane, and one variant where features are projected to all three canonical planes.
While the former is computationally more efficient, the latter allows for recovering richer geometric structure in the $z$ dimension.

\boldparagraph{Volume Encoder}
While planar feature representations allow for encoding at large spatial resolution ($128^2$ pixels and beyond), they are restricted to two dimensions. Therefore, we also consider volumetric encodings (see \figref{fig:model_b}) which better represent 3D information, but are restricted to smaller resolutions (typically $32^3$ voxels in our experiments).
Similar to the plane encoder, we perform average pooling, but this time over all features falling into the same \textit{voxel} cell, resulting in a feature volume of dimensionality $H \times W \times D \times d$.

\subsection{Decoder}\label{subsec:decoder}
We endow our model with translation equivariance by processing the feature planes and the feature volume from the encoder using 2D and 3D convolutional hourglass (U-Net) networks \cite{Ronneberger2015MICCAI,Cciccek2016MICCAI}
which are composed of a series of down- and upsampling convolutions with skip connections to integrate both local and global information.
We choose the depth of the U-Net such that the receptive field becomes equal to the size of the respective feature plane or volume.

Our single-plane decoder (\figref{fig:model_c}) processes the ground plane features with a 2D U-Net.
The multi-plane decoder (\figref{fig:model_d}) processes each feature plane separately using 2D U-Nets with shared weights.
Our volume decoder (\figref{fig:model_e}) uses a 3D U-Net.
Since convolution operations are translational equivariant, our output features are also translation equivariant, enabling structured reasoning.
Moreover, convolutional operations are able to ``inpaint'' features while preserving global information, enabling reconstruction from sparse inputs.

\subsection{Occupancy Prediction}
\label{subsec:occupancy-prediction}

Given the aggregated feature maps, our goal is to estimate the occupancy probability of any point $\bp$ in 3D space.
For the single-plane decoder, we project each point $\bp$ orthographically onto the ground plane and query the feature value through bilinear interpolation (\figref{fig:model_c}).
For the multi-plane decoder (\figref{fig:model_d}), we aggregate information from the 3 canonical planes by summing the features of all 3 planes.
For the volume decoder, we use trilinear interpolation (\figref{fig:model_e}).

Denoting the feature vector for input $\bx$ at point $\bp$ as $\psi(\bp, \bx)$, we predict the occupancy of $\bp$ using a small fully-connected occupancy network:
\begin{equation}
  f_{\theta}(\bp, \psi(\bp, \bx)) \rightarrow [0, 1]
\end{equation}
The network comprises multiple ResNet blocks.
We use the network architecture of \cite{Niemeyer2019ARXIV}, adding $\psi$ to the input features of every ResNet block instead of the more memory intensive batch normalization operation proposed in earlier works \cite{Mescheder2019CVPR}. In contrast to \cite{Niemeyer2019ARXIV}, we use a feature dimension of $32$ for the hidden layers. Details about the network architecture can be found in the supplementary.

\subsection{Training and Inference}
\label{subsec:training}

At training time, we uniformly sample query points $\bp\in\nR^3$ within the volume of interest and predict their occupancy values.
We apply the binary cross-entropy loss between the predicted $\hat{o}_{\bp}$ and the true occupancy values $o_{\bp}$:
\begin{equation}
  \cL(\hat{o}_{\bp}, o_{\bp}) = - [o_{\bp}\cdot \log (\hat{o}_{\bp}) + (1 - o_{\bp})\cdot \log(1-\hat{o}_{\bp})]
  \label{eq:loss_cross_entropy}
\end{equation}
We implement all models in PyTorch~\cite{Pytorch2019NIPS} and use the Adam~optimizer~\cite{Kingma2015ICML} with a learning rate of $10^{-4}$.
During inference, we apply \emph{Multiresolution IsoSurface Extraction} (MISE)~\cite{Mescheder2019CVPR} to extract meshes given an input $\bx$.
As our model is fully-convolutional, we are able to reconstruct large scenes by applying it in a ``sliding-window'' fashion at inference time. We exploit this property to obtain reconstructions of entire apartments (see \figref{fig:teaser}).

\section{Experiments}\label{sec:experiments}

We conduct three types of experiments to evaluate our method. %
First, we perform \textbf{object-level reconstruction} on ShapeNet~\cite{Chang2015ARXIV} chairs, considering noisy point clouds and low-resolution occupancy grids as inputs.
Next, we compare our approach against several baselines on the task of \textbf{scene-level reconstruction} using a synthetic indoor dataset of various objects.
Finally, we demonstrate \textbf{synthetic-to-real generalization} by evaluating our model on real indoor scenes~\cite{Dai2017CVPR, Chang2017THREEDV}. %

\boldparagraph{\underline{Datasets}}\vspace{0.2cm}

\noindent\textbf{ShapeNet~\cite{Chang2015ARXIV}:} We use all 13 classes of the ShapeNet subset, voxelizations, and train/val/test split from Choy~\etal\cite{Choy2016ECCV}. Per-class results can be found in supplementary.

\noindent\textbf{Synthetic Indoor Scene Dataset:} We create a synthetic dataset of 5000 scenes with multiple objects from ShapeNet (chair, sofa, lamp, cabinet, table).
A scene consists of a ground plane with randomly sampled width-length ratio, multiple objects with random rotation and scale, and randomly sampled walls.

\noindent\textbf{ScanNet v2~\cite{Dai2017CVPR}:} This dataset contains 1513 real-world rooms captured with an RGB-D camera. We sample point clouds from the provided meshes for testing.

\noindent\textbf{Matterport3D~\cite{Chang2017THREEDV}:} Matterport3D contains 90 buildings with multiple rooms on different floors captured using a Matterport Pro Camera. 
Similar to ScanNet, we sample point clouds for evaluating our model on Matterport3D.

\boldparagraph{\underline{Baselines}}\vspace{0.2cm}

	\noindent\textbf{ONet~\cite{Mescheder2019CVPR}:} Occupancy Networks is a state-of-the-art implicit 3D reconstruction model. It uses a fully-connected network architecture and a global encoding of the input. We compare against this method in all of our experiments.

\noindent\textbf{PointConv:} We construct another simple baseline by extracting point-wise features using PointNet++~\cite{Qi2017NIPS}, interpolating them using Gaussian kernel regression and feeding them into the same fully-connected network used in our approach.
While this baseline uses local information, it does not exploit convolutions.

\noindent\textbf{SPSR~\cite{Kazhdan2013SIGGRAPH}:} Screened Poisson Surface Reconstruction (SPSR) is a traditional 3D reconstruction technique which operates on oriented point clouds as input. 
Note that in contrast to all other methods, SPSR requires additional surface normals which are often hard to obtain for real-world scenarios.

\clearpage %
\boldparagraph{\underline{Metrics}}\vspace{0.2cm}

\noindent Following~\cite{Mescheder2019CVPR}, we consider Volumetric IoU, Chamfer Distance, Normal Consistency for evaluation. We further report F-Score~\cite{Tatarchenko2019CVPR} with the default threshold value of 1\% unless otherwise specified.
Details can be found in the supplementary.

\subsection{Object-Level Reconstruction}
We first evaluate our method on the single object reconstruction task on ShapeNet~\cite{Chang2015ARXIV}.
We consider two different types of 3D inputs: noisy point clouds and low-resolution voxels.
For the former, we sample $3000$ points from the mesh and apply Gaussian noise with zero mean and standard deviation $0.05$.
For the latter, we use the coarse $32^3$ voxelizations from~\cite{Mescheder2019CVPR}.
For the query points (\ie, for which supervision is provided), we follow~\cite{Mescheder2019CVPR} and uniformly sample 2048 and 1024 points for noisy point clouds and low-resolution voxels, respectively. 
Due to the different encoder architectures for these two tasks, we set the batch size to 32 and 64, respectively.

\begin{figure}[!bt]
	\centering
	\newcommand{\mywidth}{0.12\textwidth}
	\begin{tabular}{c|ccccc|c}
		\includegraphics[width=\mywidth,trim={9cm 7cm 9cm 7cm}, clip]{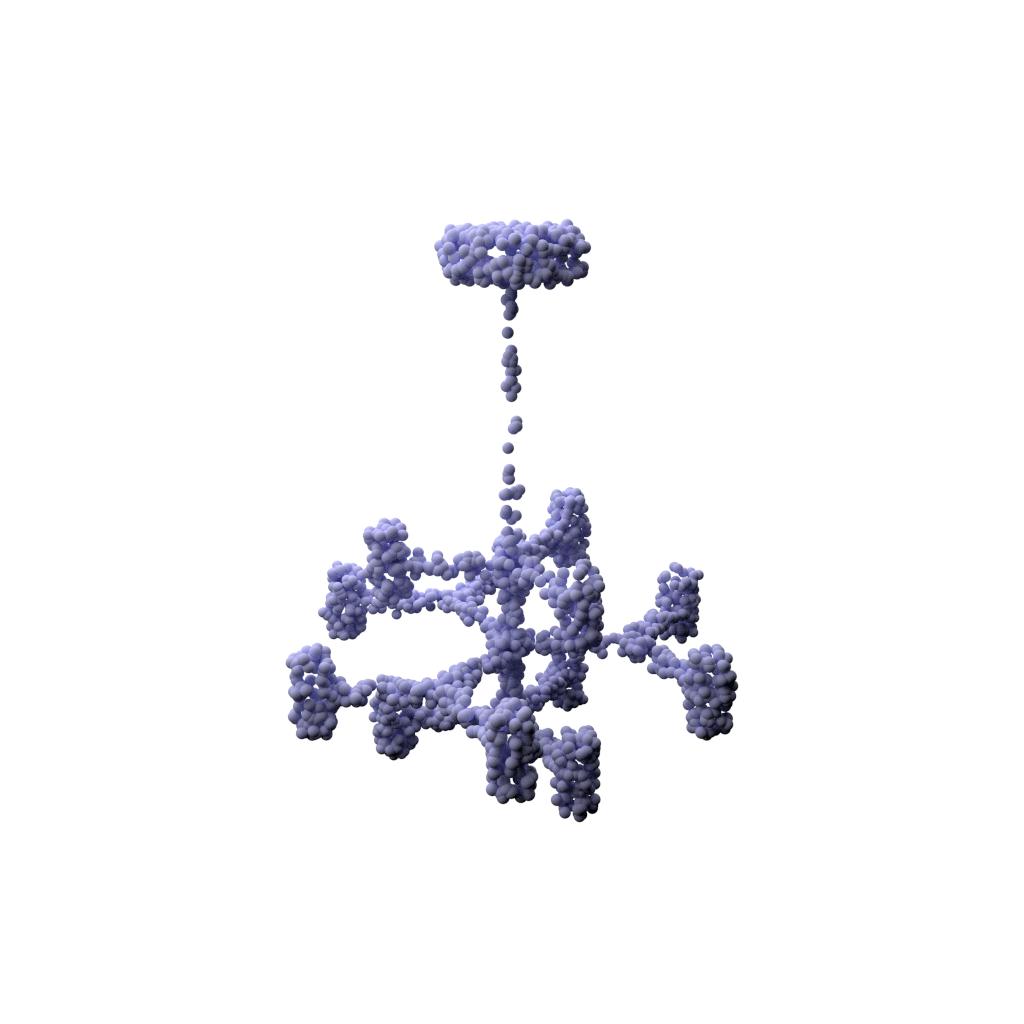}~~&~
		\includegraphics[width=\mywidth,trim={9cm 7cm 9cm 7cm}, clip]{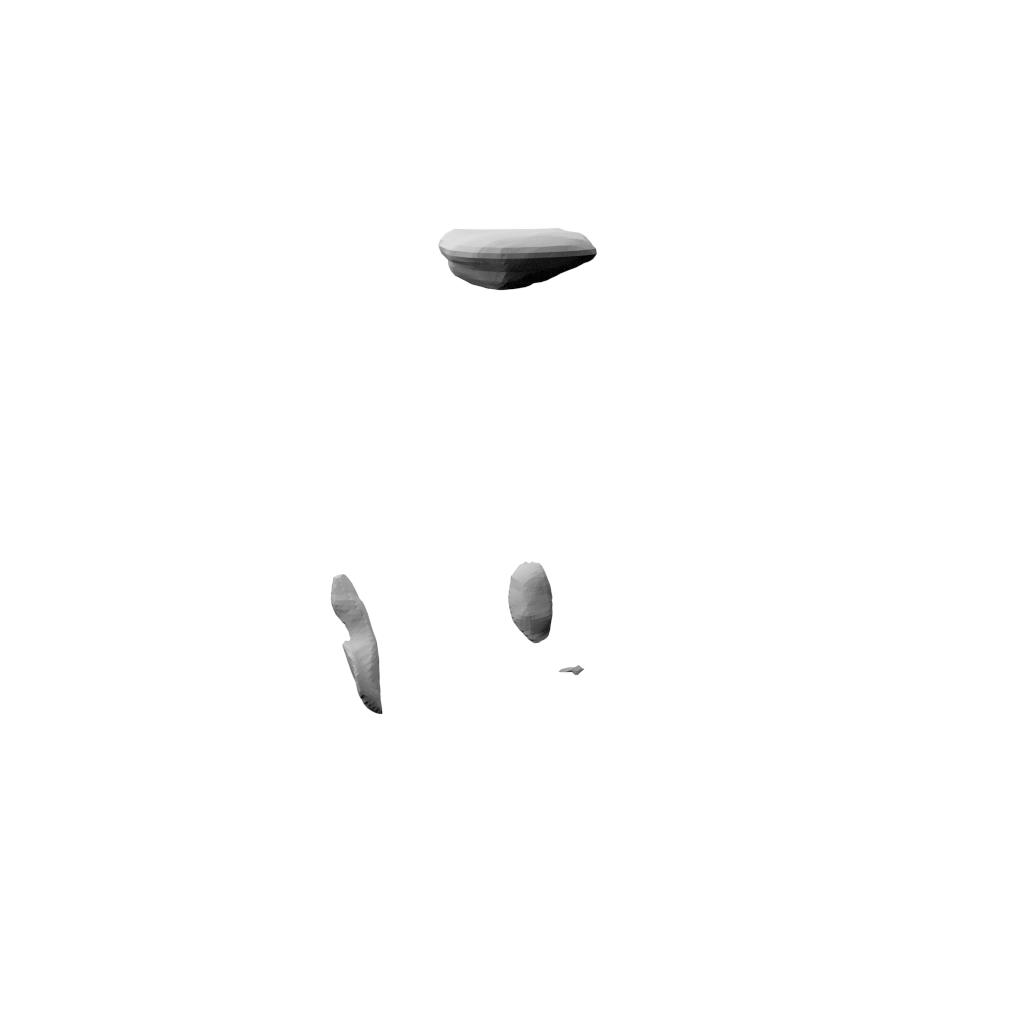} & 		
		\includegraphics[width=\mywidth,trim={9cm 7cm 9cm 7cm}, clip]{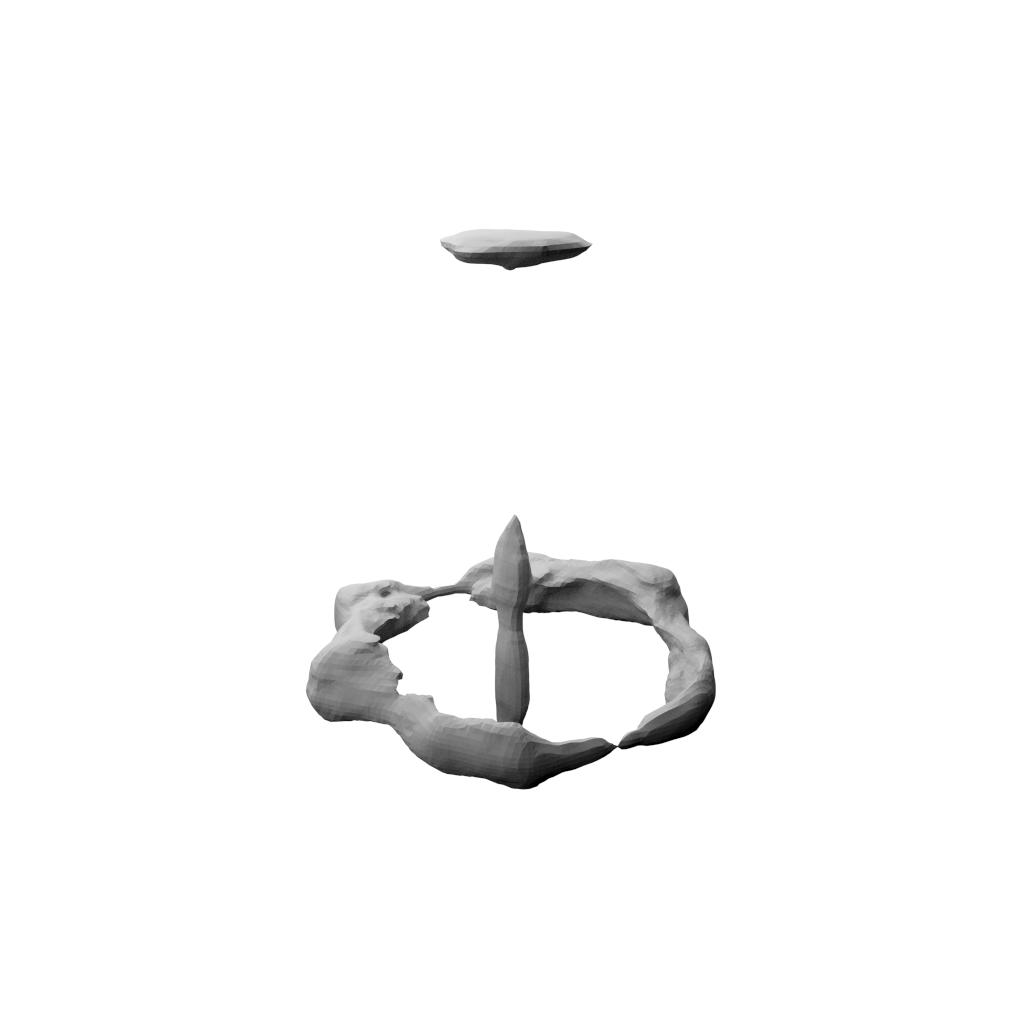} &
		\includegraphics[width=\mywidth,trim={9cm 7cm 9cm 7cm}, clip]{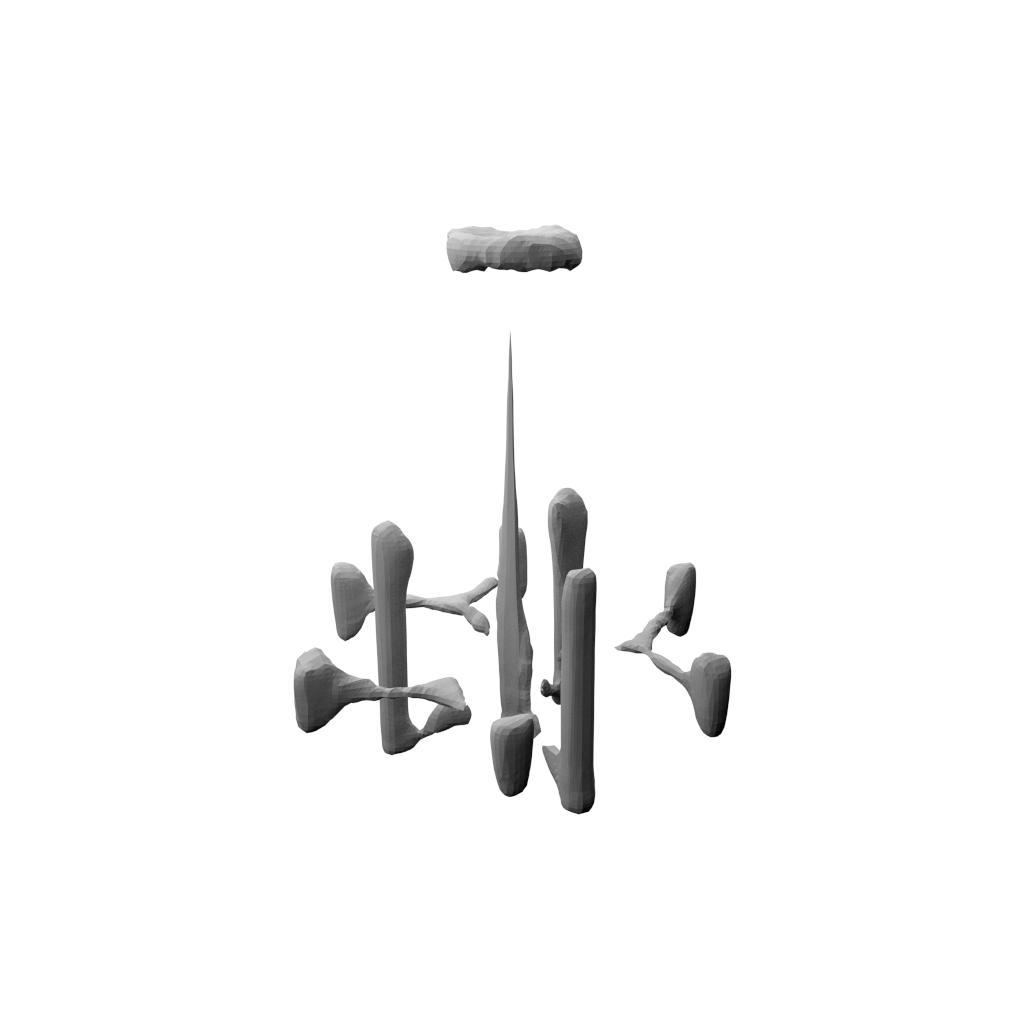}&
		\includegraphics[width=\mywidth,trim={9cm 7cm 9cm 7cm}, clip]{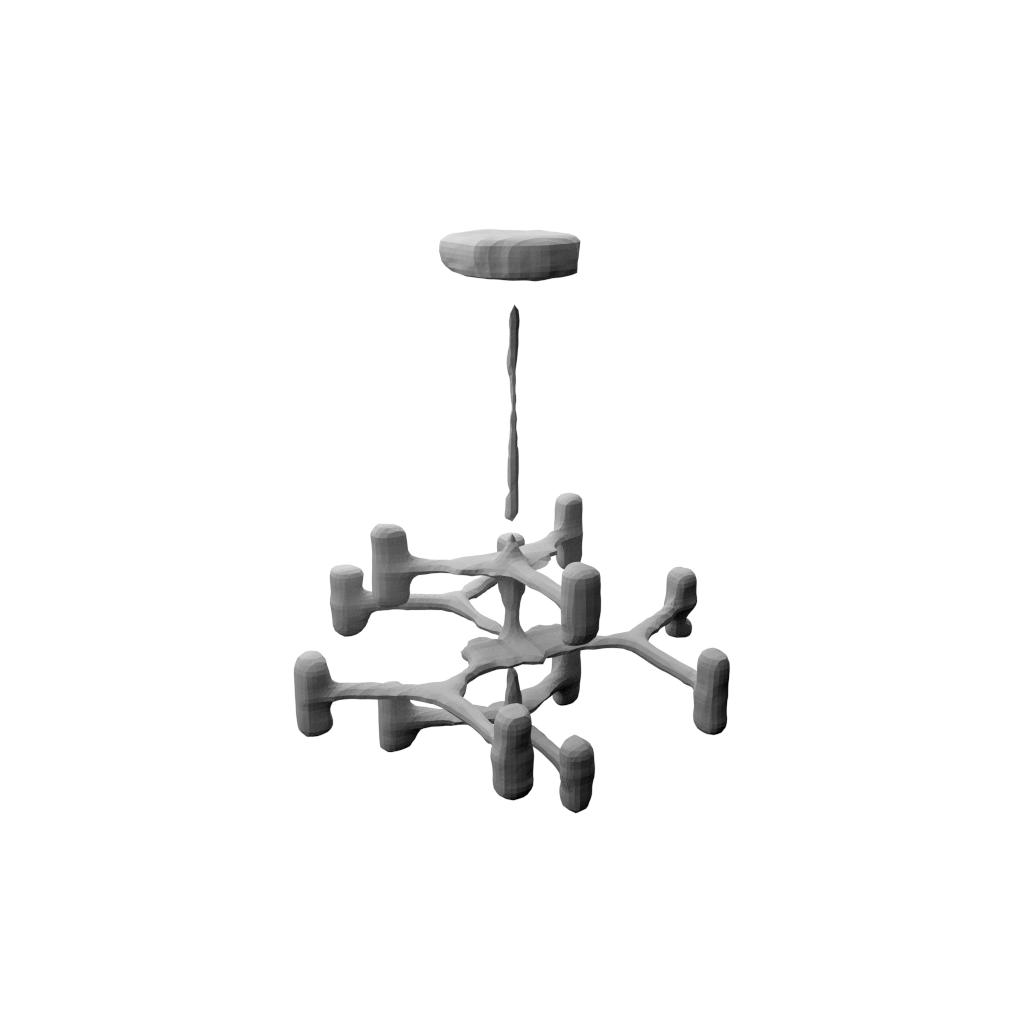}&
		\includegraphics[width=\mywidth,trim={9cm 7cm 9cm 7cm}, clip]{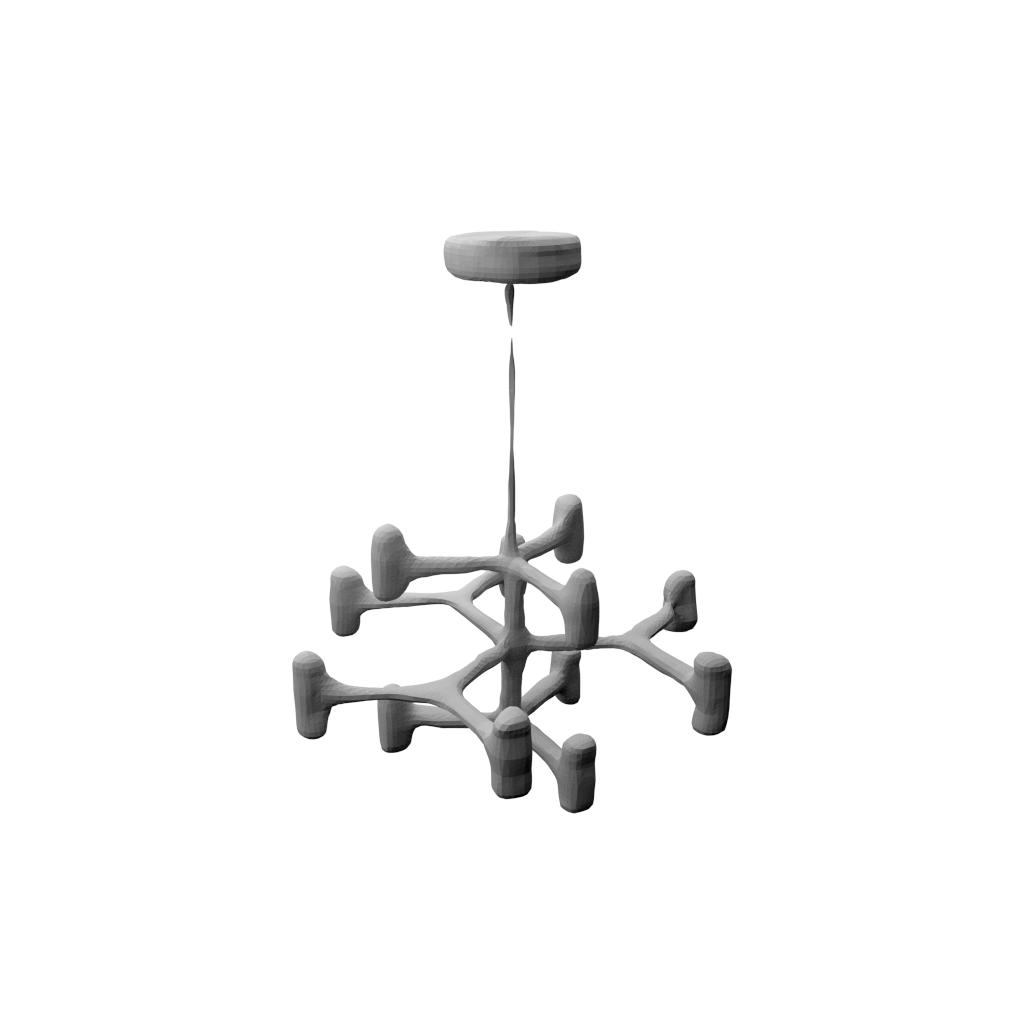} ~~&~
		\includegraphics[width=\mywidth,trim={9cm 7cm 9cm 7cm}, clip]{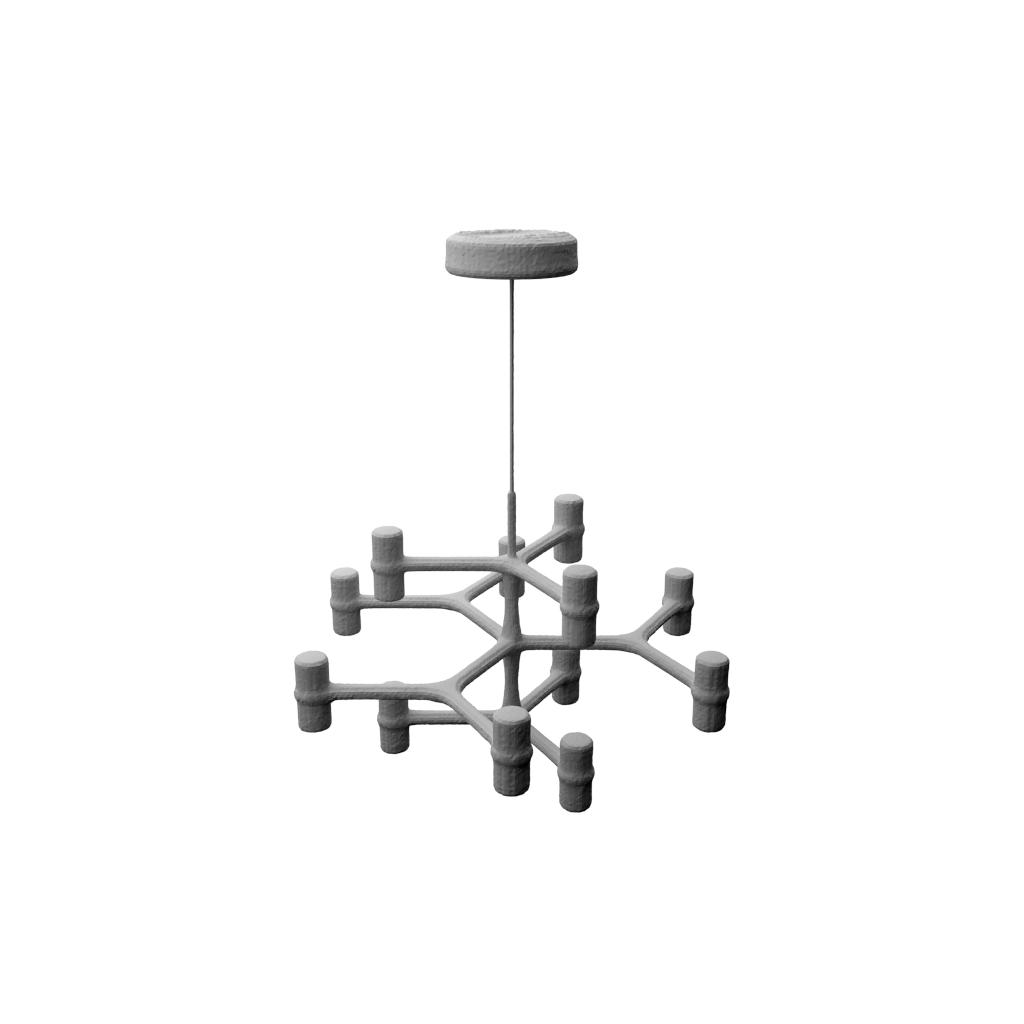}
		\\
		\includegraphics[width=\mywidth]{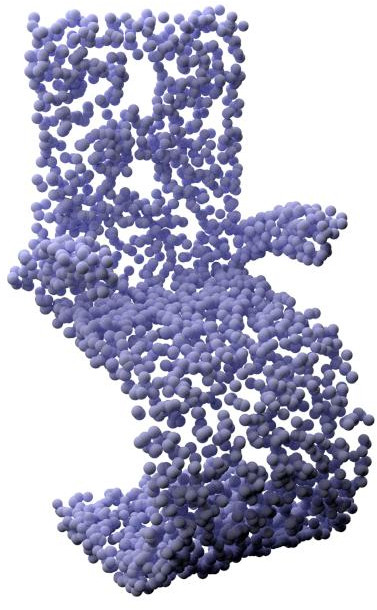}~~&~
		\includegraphics[width=\mywidth]{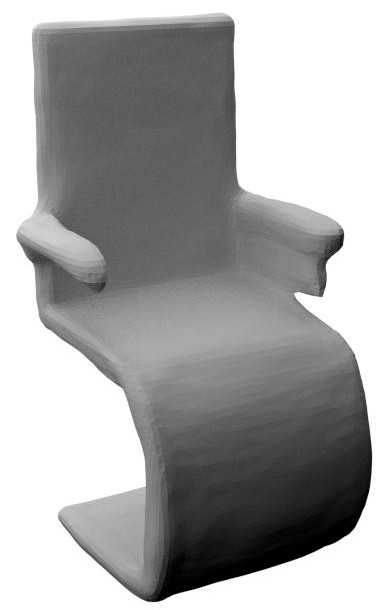} & 		
		\includegraphics[width=\mywidth]{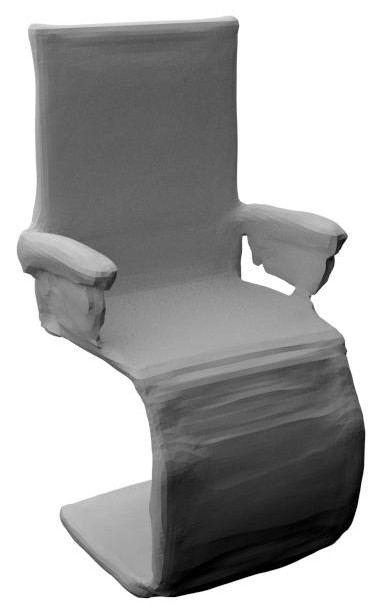} &
		\includegraphics[width=\mywidth]{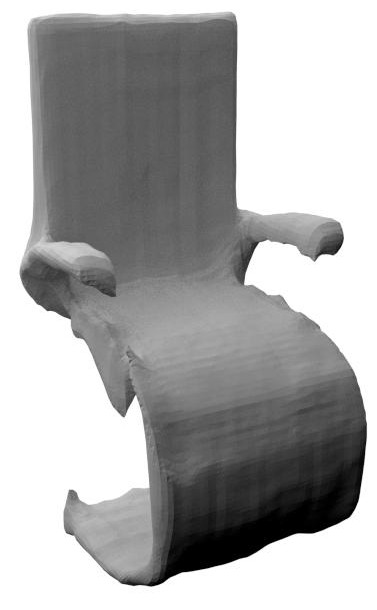}&
		\includegraphics[width=\mywidth]{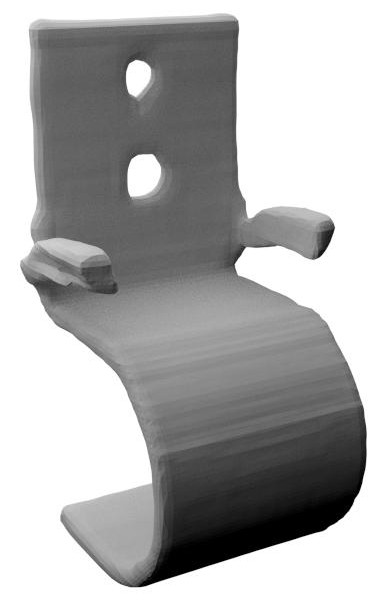}&
		\includegraphics[width=\mywidth]{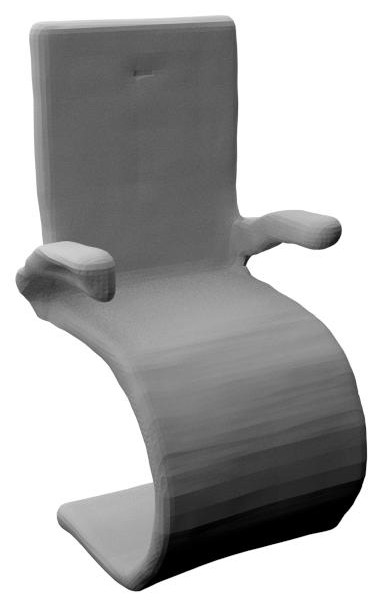} ~~&~
		\includegraphics[width=\mywidth]{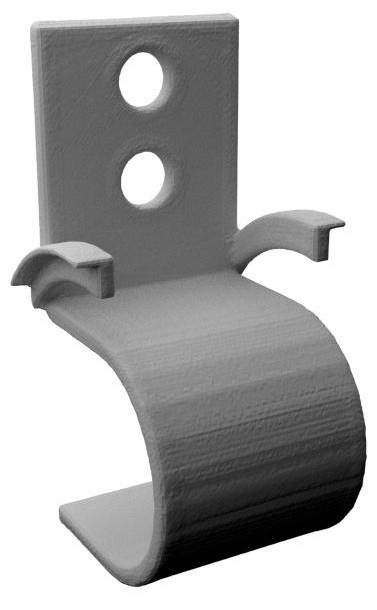}
		\\		
		\includegraphics[width=\mywidth,trim={8cm 10cm 9cm 10cm}, clip]{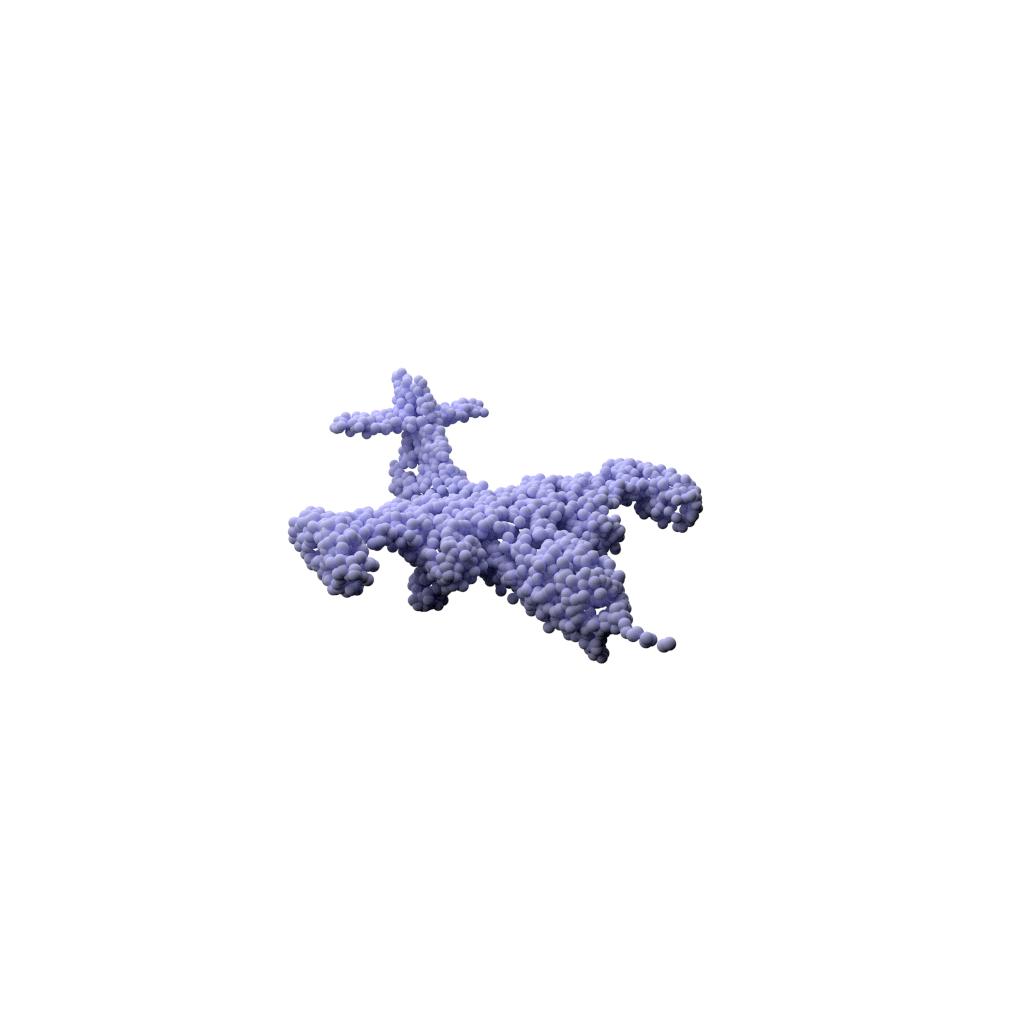}~~&~
		\includegraphics[width=\mywidth,trim={8cm 10cm 9cm 10cm}, clip]{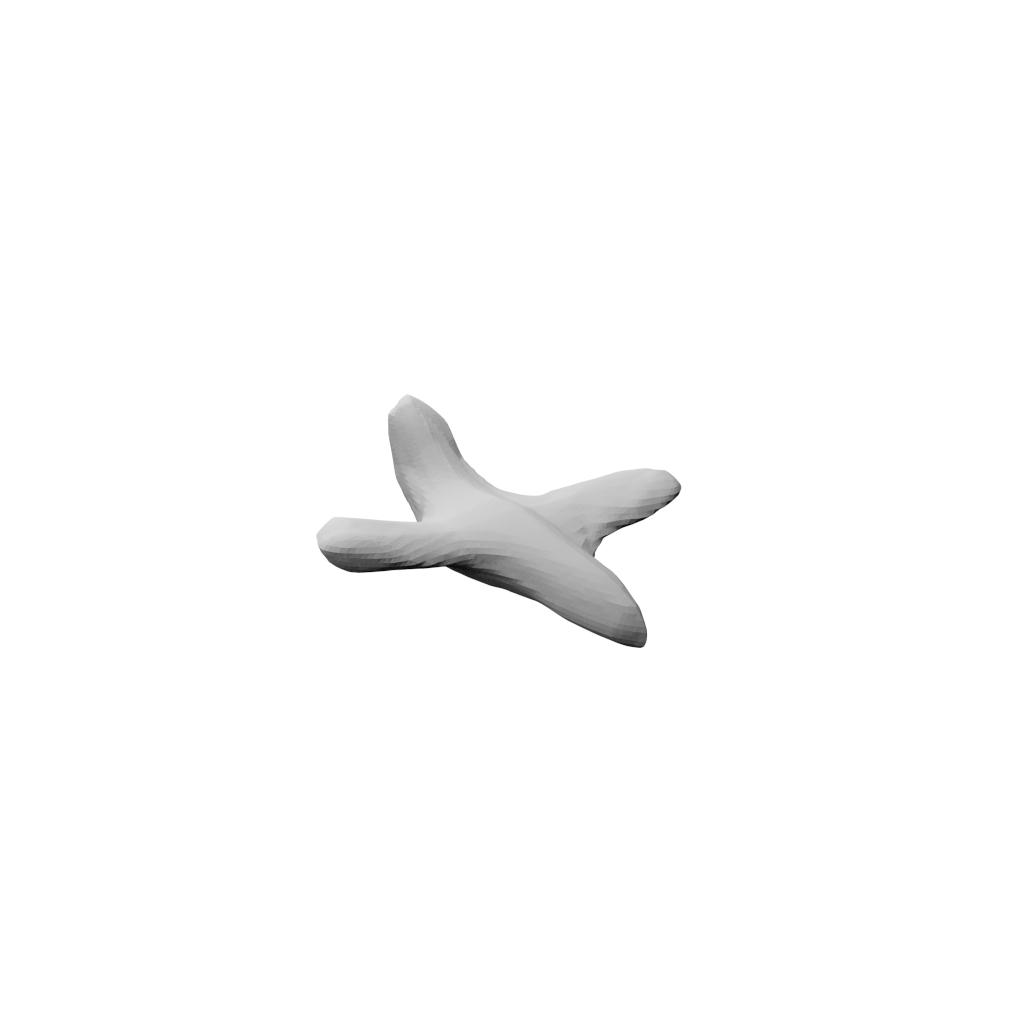} & 		
		\includegraphics[width=\mywidth,trim={8cm 10cm 9cm 10cm}, clip]{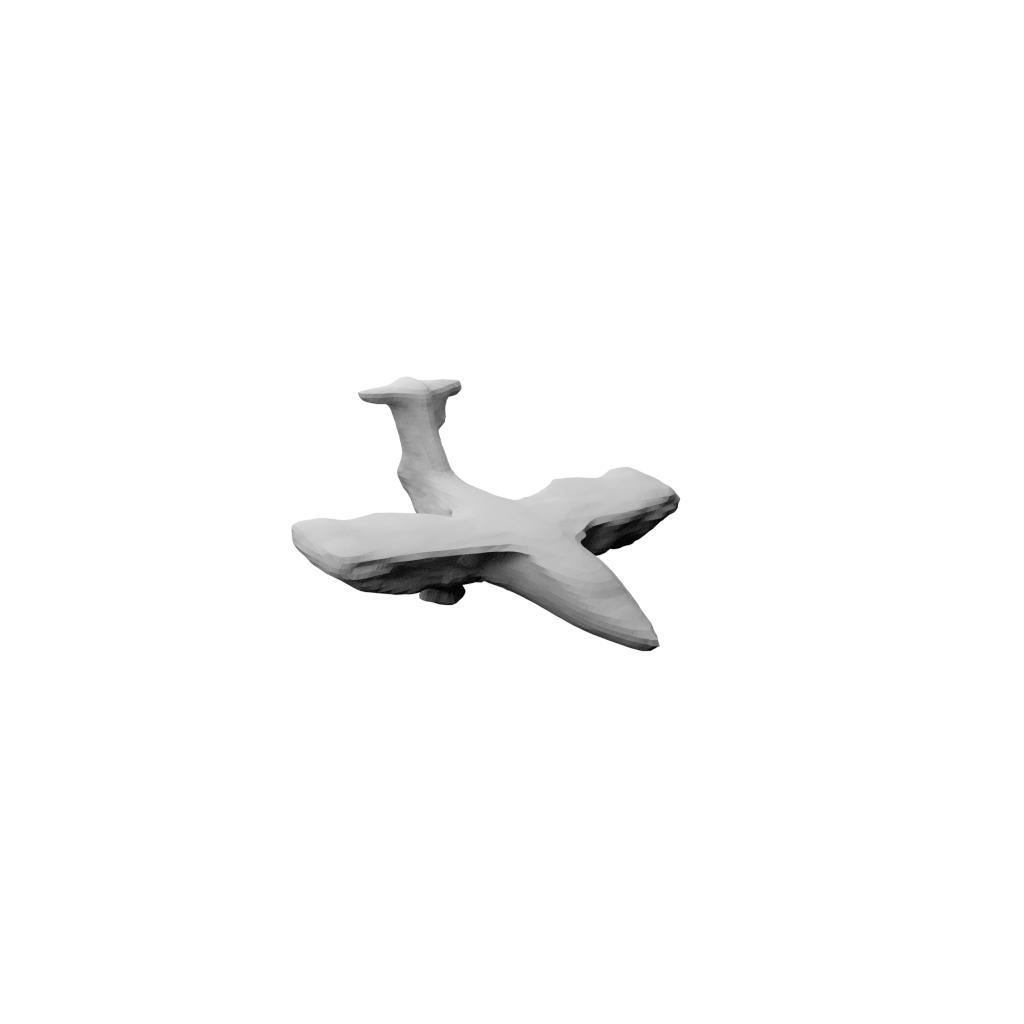} &
		\includegraphics[width=\mywidth,trim={8cm 10cm 9cm 10cm}, clip]{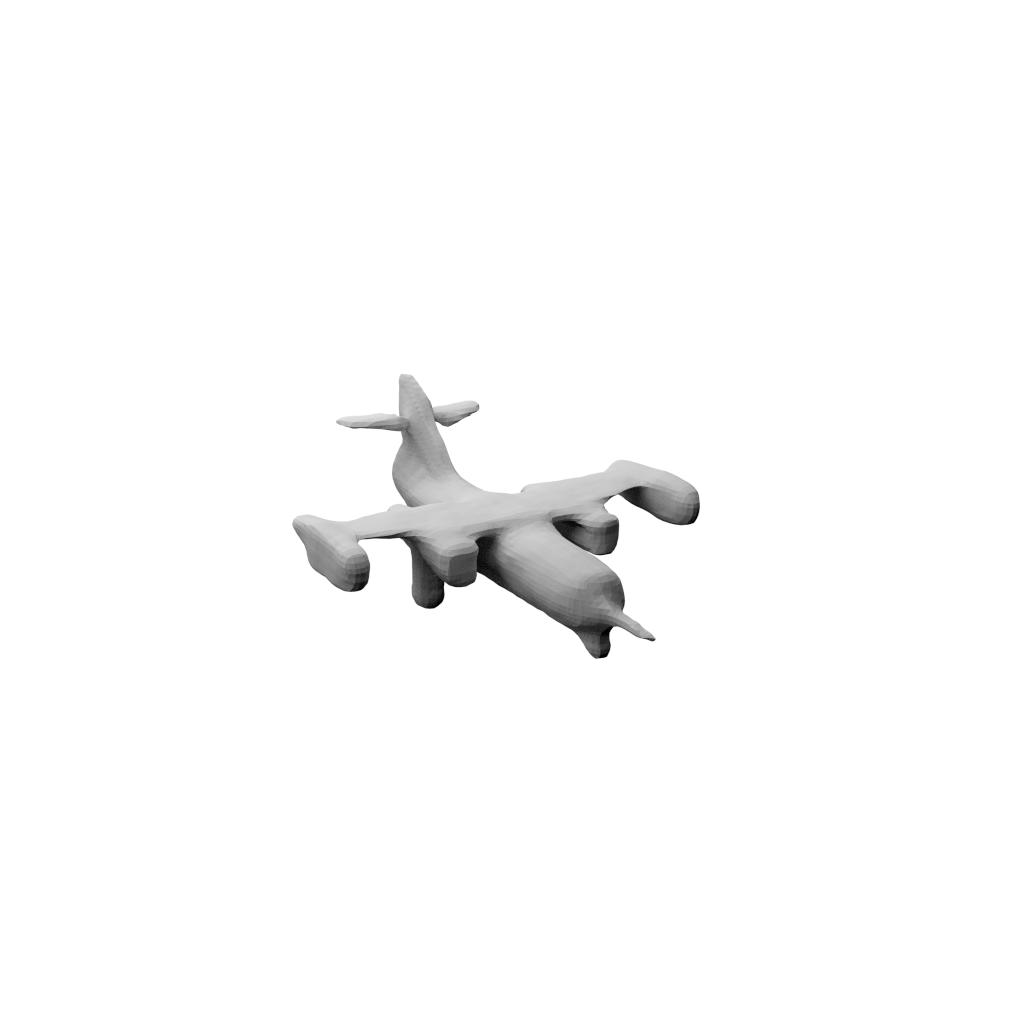}&
		\includegraphics[width=\mywidth,trim={8cm 10cm 9cm 10cm}, clip]{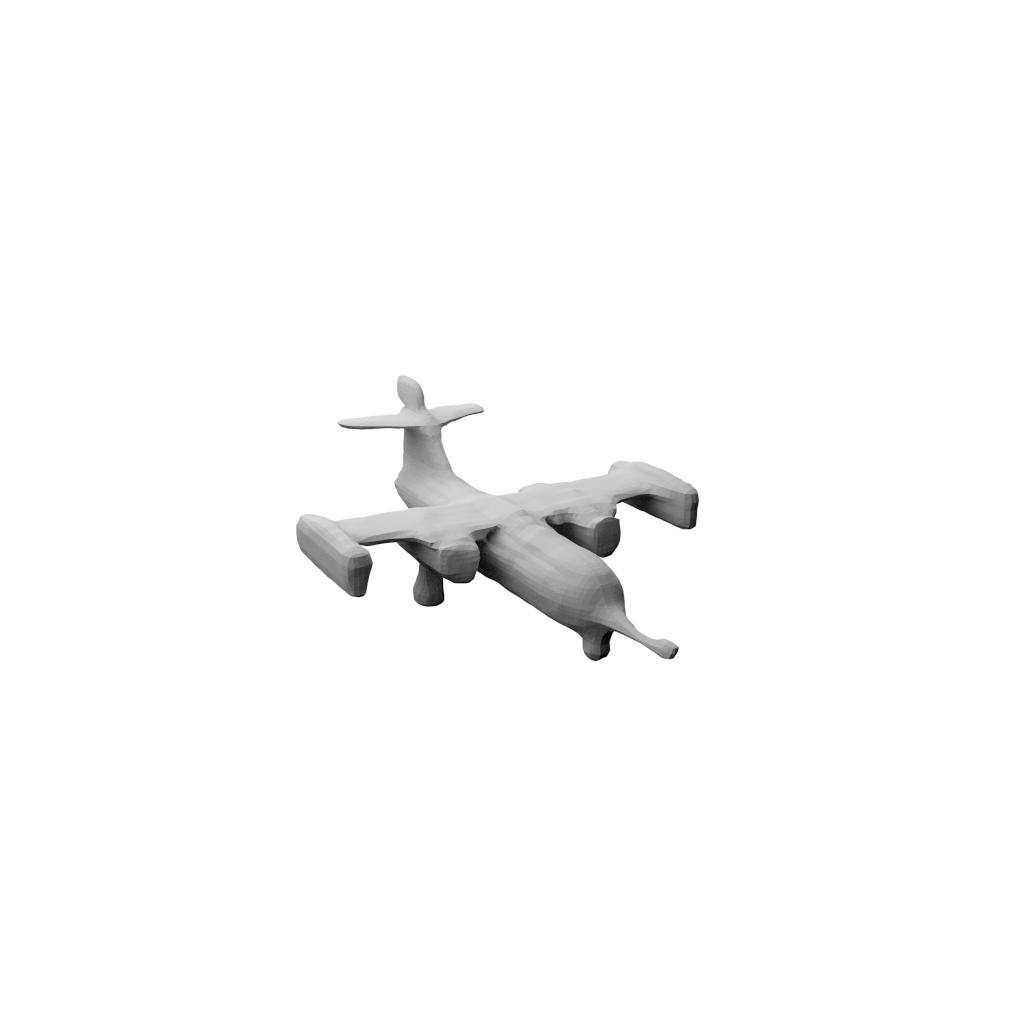}&
		\includegraphics[width=\mywidth,trim={8cm 10cm 9cm 10cm}, clip]{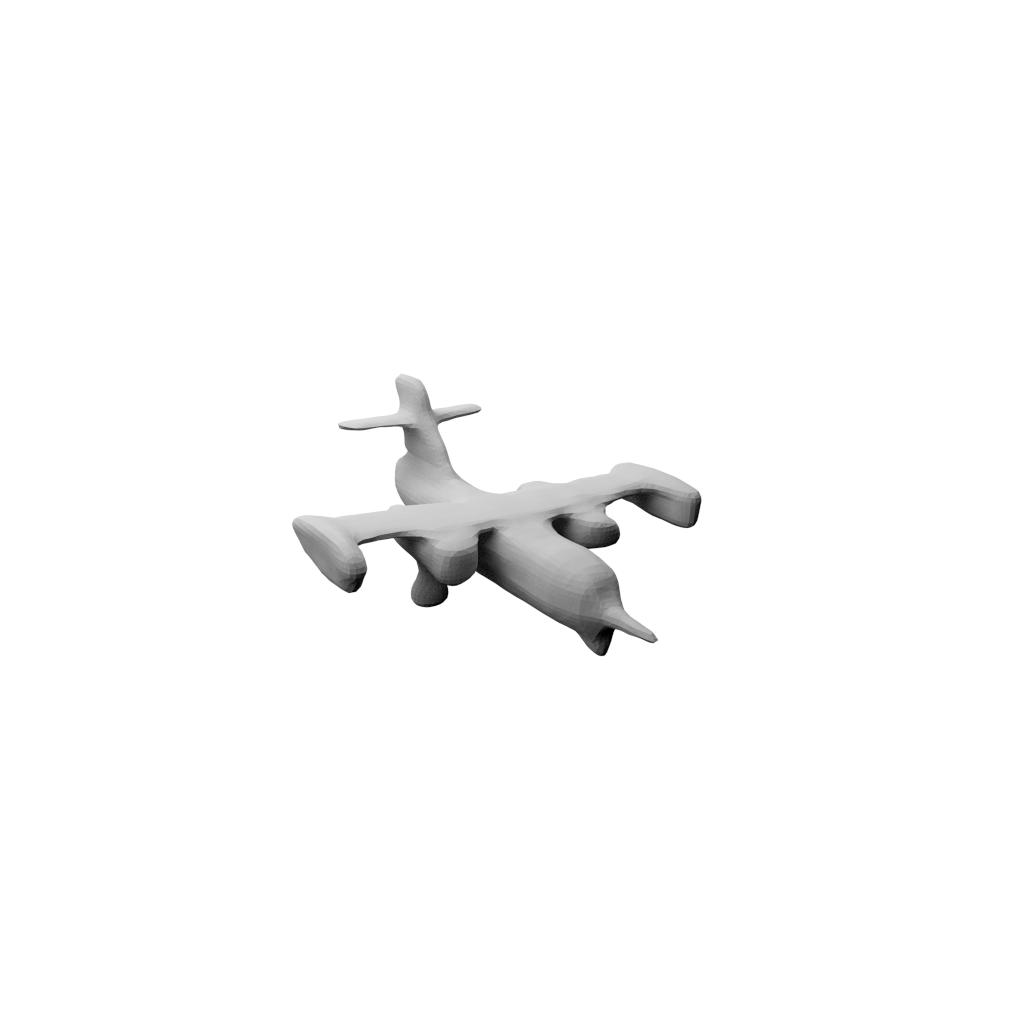} ~~&~
		\includegraphics[width=\mywidth,trim={8cm 10cm 9cm 10cm}, clip]{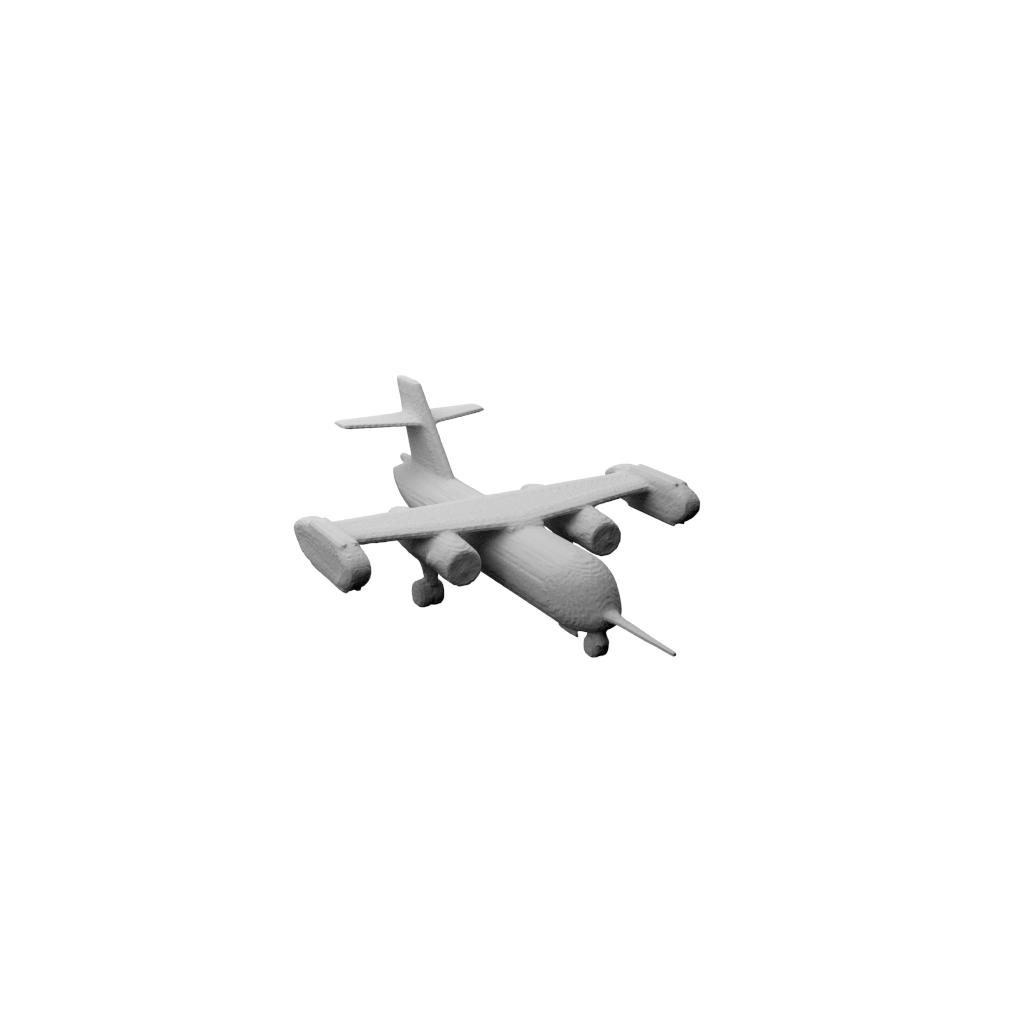}
		\\
		Input & PointConv& ONet~\cite{Mescheder2019CVPR} &Ours-2D & Ours-2D & Ours-3D & GT mesh\\
		&&&($64^2$)&($3 \times 64^2$)&($32^3$) &
	\end{tabular}
	\caption{
		\textbf{Object-Level 3D Reconstruction from Point Clouds.}
	 	Comparison of our convolutional representation to ONet and PointConv on ShapeNet.
	}
	\label{fig:shapenet_chair}
\end{figure}

\boldparagraph{Reconstruction from Point Clouds}%
\tabref{tab:shapenet_chair} and \figref{fig:shapenet_chair} show quantitative and qualitative results.
Compared to the baselines, all variants of our method achieve equal or better results on all three metrics.
As evidenced by the training progression plot on the right, our method reaches a high validation IoU after only few iterations.
This verifies our hypothesis that leveraging convolutions and local features benefits 3D reconstruction in terms of both accuracy and efficiency.
The results show that, in comparison to PointConv which directly aggregates features from point clouds, projecting point-features to planes or volumes followed by 2D/3D CNNs is more effective.
In addition, decomposing 3D representations from volumes into three planes with higher resolution ($64^2$ vs. $32^3$) improves performance while at the same time requiring less GPU memory.
More results can be found in supplementary.

\begin{table*}[!bt]
	\centering
	\setlength{\tabcolsep}{0.2cm}
	\begin{tabular}{cc}
		\resizebox{0.57\textwidth}{!}{
			\begin{tabular}{lccccc}
    \toprule
    {} &         GPU Memory   &  IoU &   Chamfer-$L_1$ & Normal C. & F-Score \\
    \midrule
    PointConv  & 5.1G&0.689& 0.126&0.858 &0.644\\
    ONet~\cite{Mescheder2019CVPR}     & 7.7G&  0.761 &  0.087 &              0.891  &0.785\\\hline
    Ours-2D ($64^2$)&     1.6G     & 0.833 &           0.059 &              0.914 & 0.887\\
    Ours-2D ($3\times64^2$)&  2.4G &         \textbf{0.884} &           \textbf{0.044} &              \textbf{0.938} & \textbf{0.942}\\
    Ours-3D ($32^3$)        & 5.9G         &  0.870 & 0.048 & 0.937 & 0.933\\
    \bottomrule
    \end{tabular}}&
		\begin{minipage}{1\textwidth}
			\includegraphics[width=0.38\textwidth]{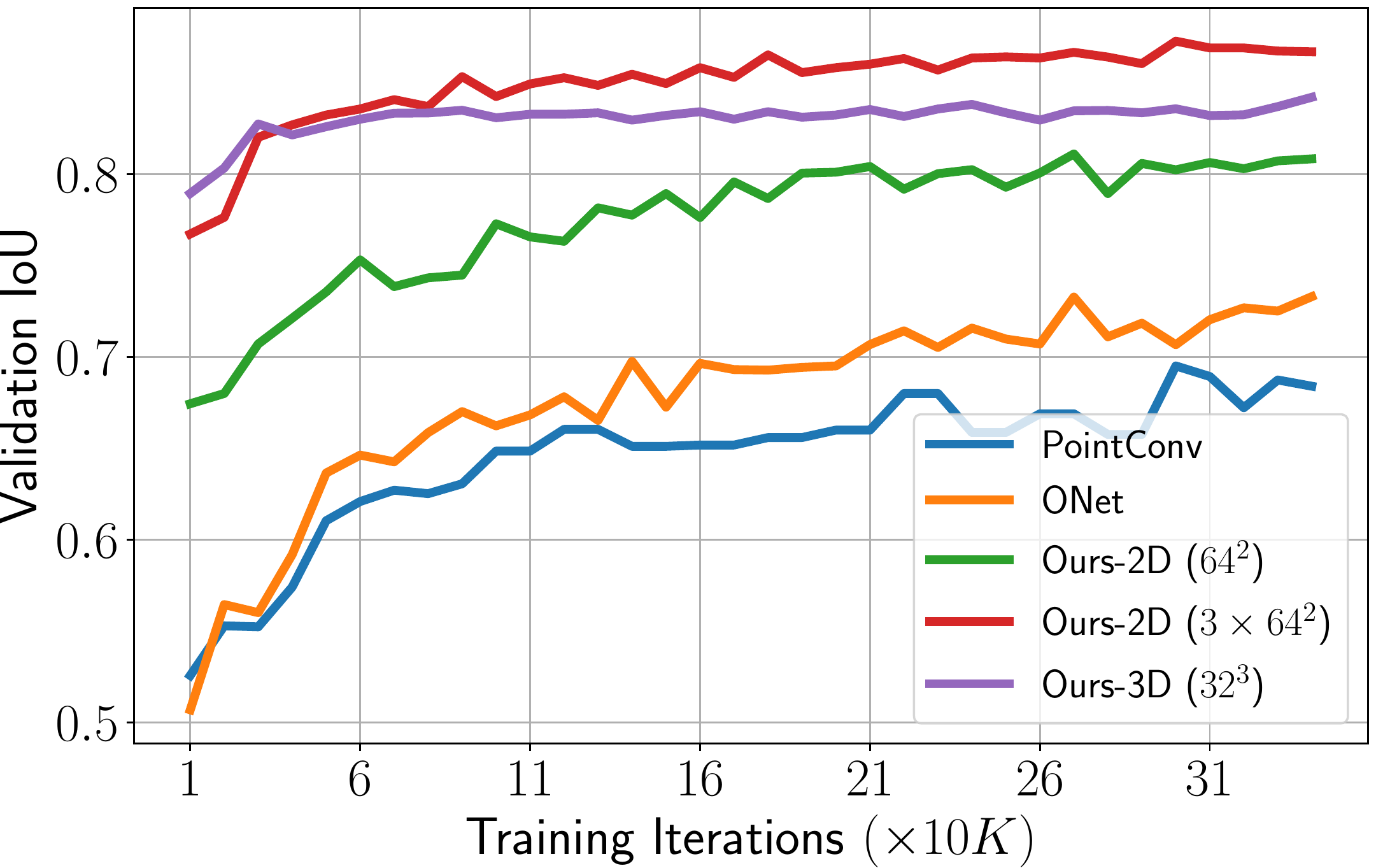}
		\end{minipage}
	\end{tabular}
	\vspace{0.1cm}
	\caption{
		\textbf{Object-Level 3D Reconstruction from Point Clouds.}
		Left: We report GPU memory, IoU, Chamfer-$L_1$ distance, Normal Consistency and F-Score for our approach (2D plane and 3D voxel grid dimensions in brackets), the baselines ONet \cite{Mescheder2019CVPR} and PointConv on ShapeNet (mean over all 13 classes).
		Right: The training progression plot shows that our method converges faster than the baselines.
	}
	\label{tab:shapenet_chair}
\end{table*}

\begin{figure}[!bt]
	\centering
	\newcommand{\mywidth}{0.15\textwidth}
	\newcommand{\mywidtht}{0.12\textwidth}
	\begin{tabular}{c|cccc|c}
		\includegraphics[width=\mywidth]{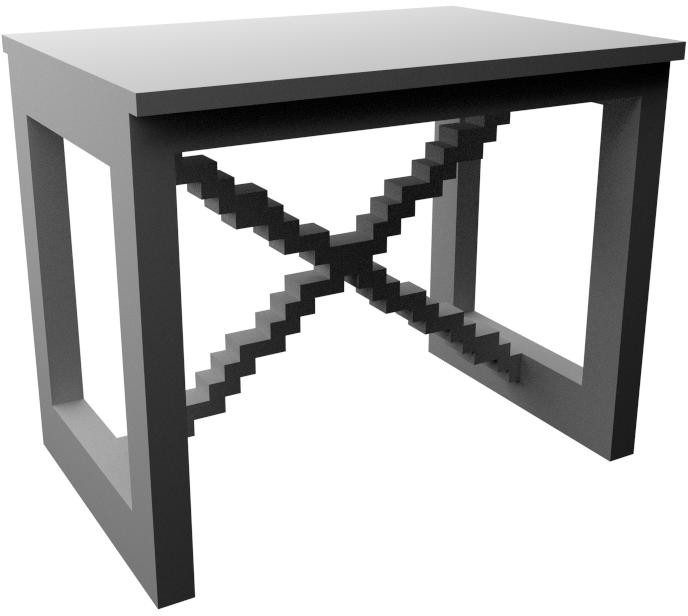} &
		\includegraphics[width=\mywidth]{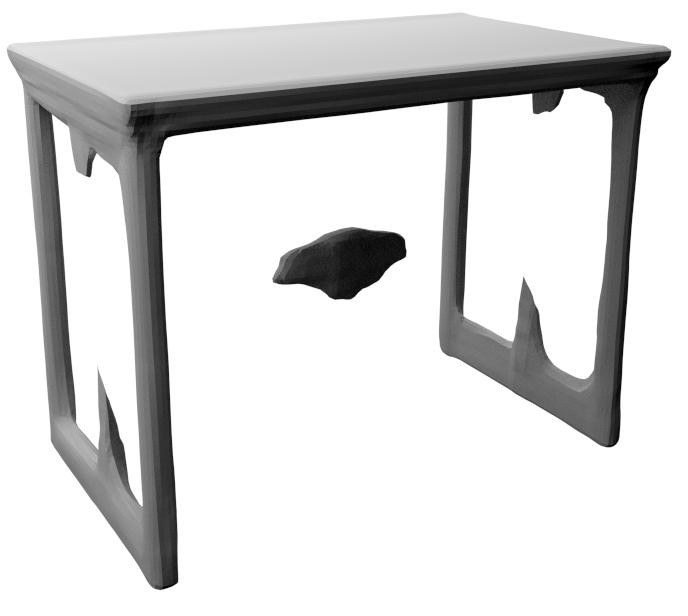} & 
		\includegraphics[width=\mywidth]{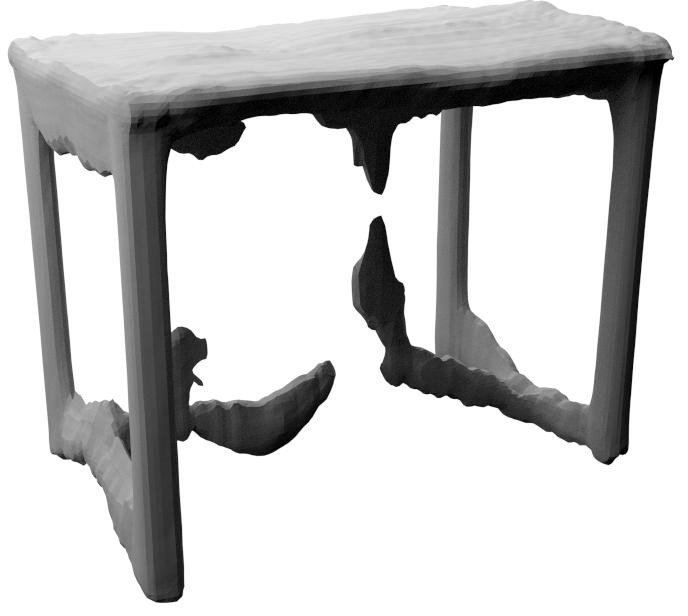} &
		\includegraphics[width=\mywidth]{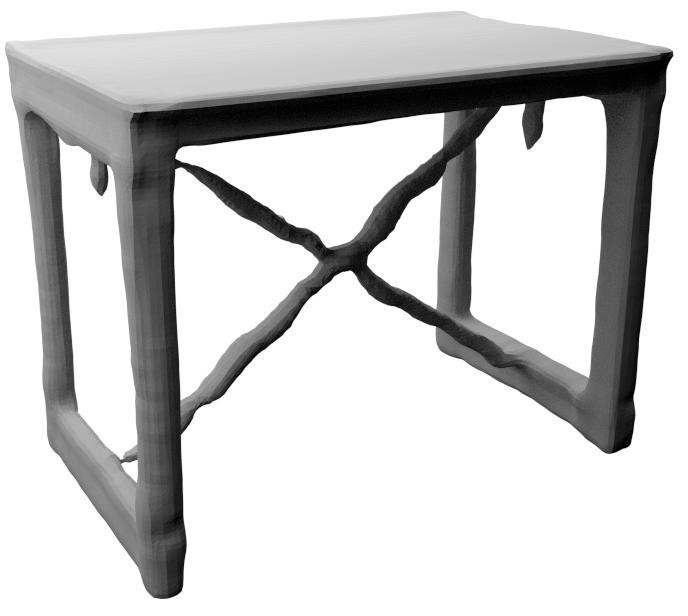}&
		\includegraphics[width=\mywidth]{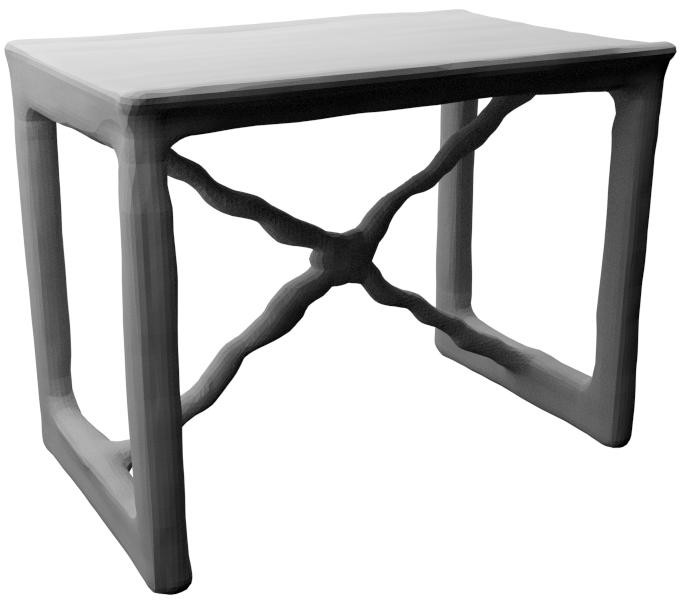}&
		\includegraphics[width=\mywidth]{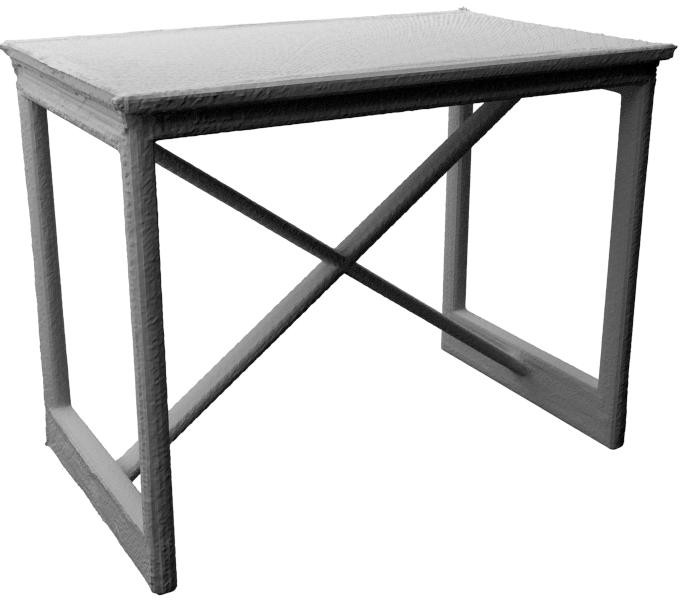}
		\\
		\includegraphics[width=\mywidtht]{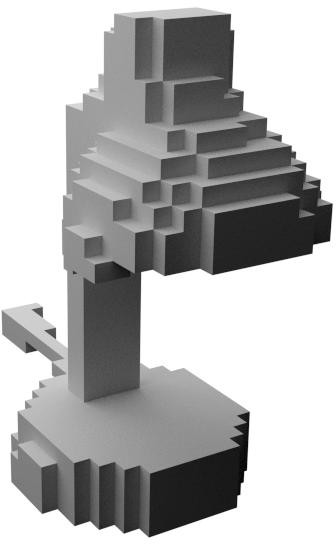} &
		\includegraphics[width=\mywidtht]{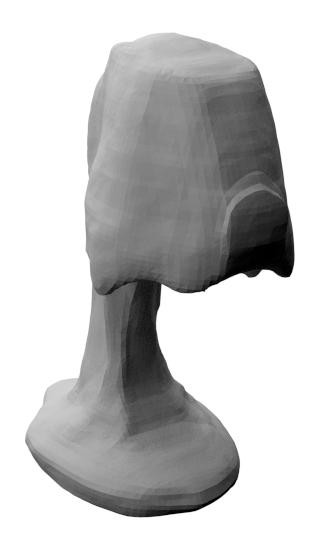} & 
		\includegraphics[width=\mywidtht]{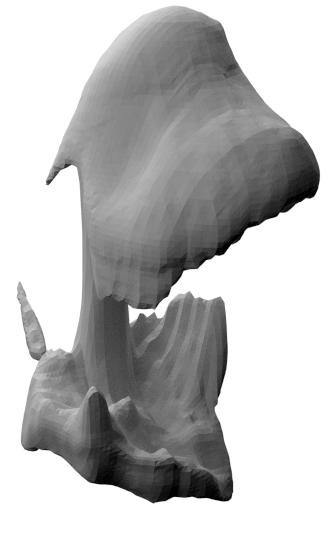} &
		\includegraphics[width=\mywidtht]{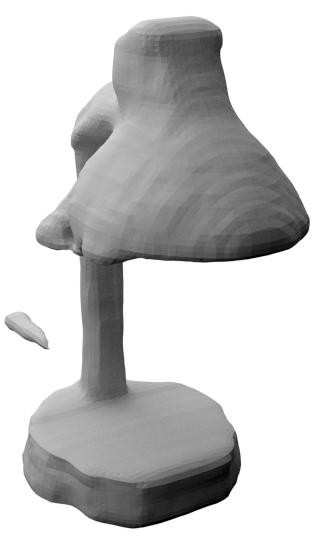}&
		\includegraphics[width=\mywidtht]{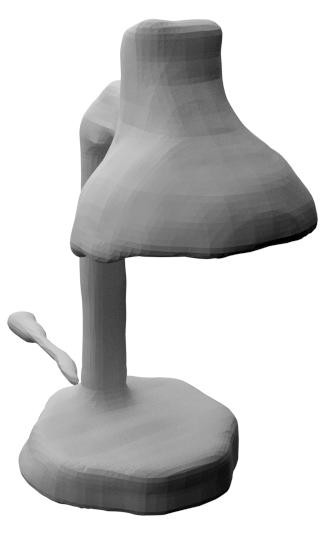}&
		\includegraphics[width=\mywidtht]{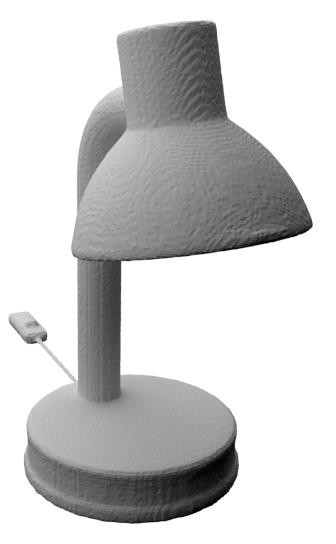}
		\\		
		Input & ONet~\cite{Mescheder2019CVPR}& Ours-2D & Ours-2D & Ours-3D  & GT mesh\\
		&   							& ($64^2$)& ($3\times64^2$) & ($32^3$) & 
		\vspace{-0.2cm}
	\end{tabular}
	\caption{
		\textbf{Voxel Super-Resolution.}
		Qualitative comparison between our method and ONet using coarse voxelized inputs at resolution $32^3$ voxels.
	}
	\label{fig:shapenet_chair_voxel}
\end{figure}

\boldparagraph{Voxel Super-Resolution}
Besides noisy point clouds, we also evaluate on the task of voxel super-resolution.
Here, the goal is to recover high-resolution details from coarse ($32^3$) voxelizations of the shape.
\tabref{tab:shapenet_chair_voxel} and \figref{fig:shapenet_chair_voxel} show that our method with three planes achieves comparable results over our volumetric method while requiring only $37\%$ of the GPU memory.
In contrast to reconstruction from point clouds, our single-plane approach fails on this task. %
We hypothesize that a single plane is not sufficient for resolving ambiguities in the coarse but regularly structured voxel input.

\begin{table*}[!bt]
	\centering
	\setlength{\tabcolsep}{0.5cm}
	\resizebox{0.9\textwidth}{!}{%
		\begin{tabular}{lccccc}
	\toprule
	{} &      GPU Memory &      IoU &   Chamfer-$L_1$ & Normal C. & F-Score\\
	\midrule
	Input & - & 0.631 & 0.136 & 0.810 & 0.440\\ \hline
	ONet~\cite{Mescheder2019CVPR} &  4.8G &  0.703 &  0.110 &    0.879 & 0.656\\\hline
	Ours-2D ($64^2$)& 2.4G  &     0.652&          0.145&    0.861        & 0.592\\
	Ours-2D ($3\times64^2$)& 4.0G  &     \textbf{0.752}&          0.092&    0.905          & \textbf{0.735}\\
	Ours-3D ($32^3$)& 10.8G &   \textbf{0.752} &         \textbf{0.091} & \textbf{0.912}        & 0.729\\
	\bottomrule
\end{tabular}
	}
	\caption{
		\textbf{Voxel Super-Resolution.}
		3D reconstruction results from low resolution voxelized inputs ($32^3$ voxels)
		on the ShapeNet dataset (mean over 13 classes). %
	}
	\label{tab:shapenet_chair_voxel}
\end{table*}

\subsection{Scene-Level Reconstruction}\label{sec:syn_indoor}
To analyze whether our approach can scale to larger scenes, we now reconstruct 3D geometry from point clouds on our synthetic indoor scene dataset.
Due to the increasing complexity of the scene, we uniformly sample $10000$ points as input point cloud and apply Gaussian noise with standard deviation of $0.05$.
During training, we sample 2048 query points, similar to object-level reconstruction.
For our plane-based methods, we use a resolution to $128^2$.
For our volumetric approach, we investigate both $32^3$ and $64^3$ resolutions.
Hypothesizing that the plane and volumetric features are complementary, we also test the combination of the multi-plane and volumetric variants.

\begin{figure}[!htb]
	\centering
	\newcommand{\mywidth}{0.185\linewidth}
	\newcommand{\mywidths}{0.22\linewidth}
	\begin{tabular}{cccccc}
		\rotatebox{90}{\qquad Input} &&
		\includegraphics[width=\mywidth]{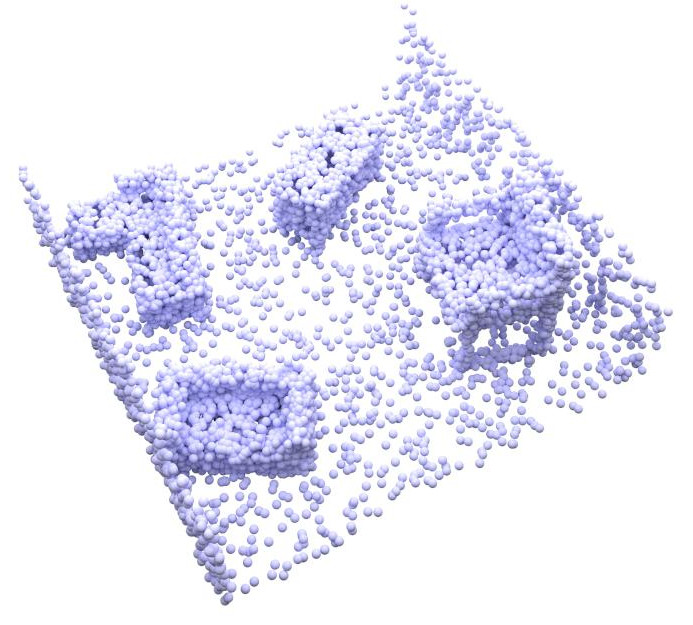} &
		\includegraphics[width=\mywidth]{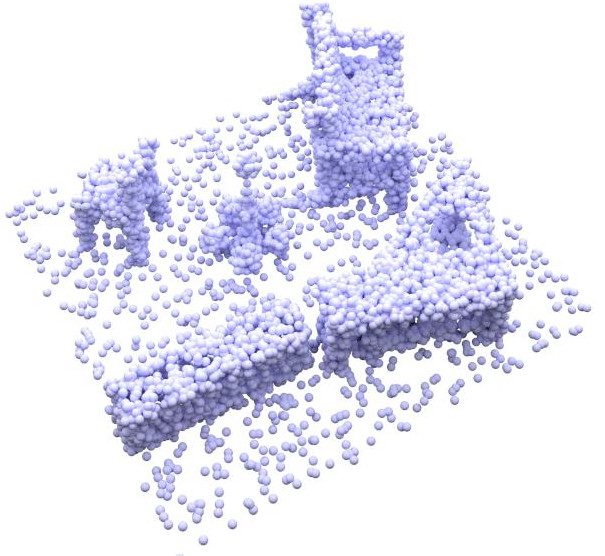} & 
		\includegraphics[width=\mywidth]{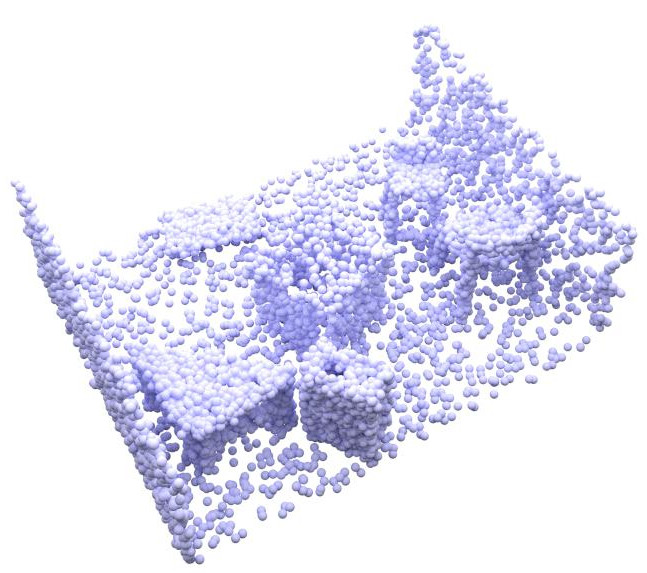} &		
		\includegraphics[width=\mywidths]{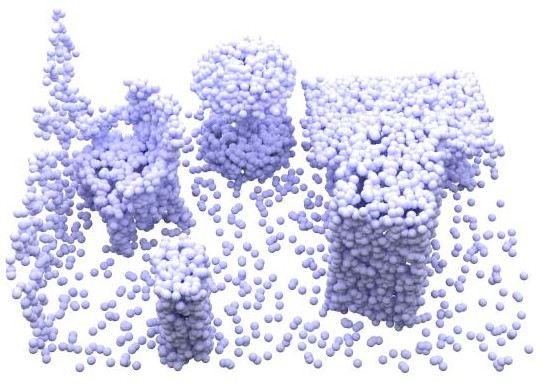}
		\\[-5pt]
		
		\rotatebox{90}{ONet~\cite{Mescheder2019CVPR}} &&
		\includegraphics[width=\mywidth]{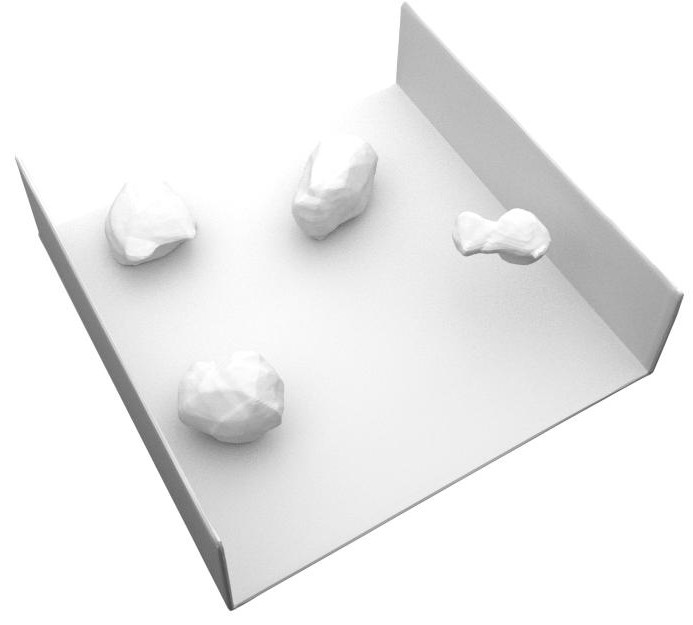} &
		\includegraphics[width=\mywidth]{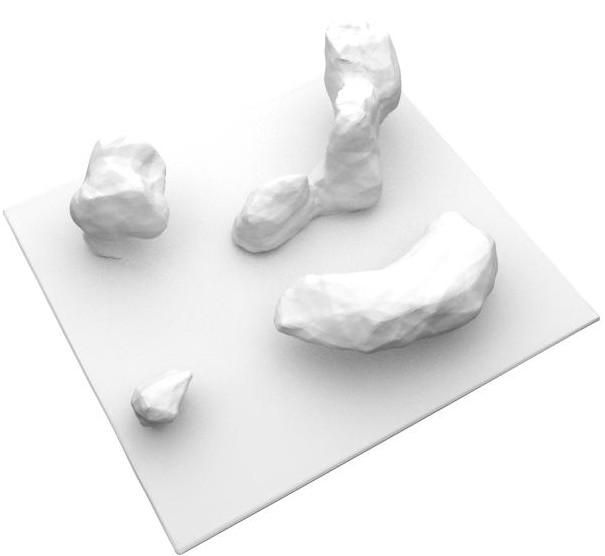} & 
		\includegraphics[width=\mywidth]{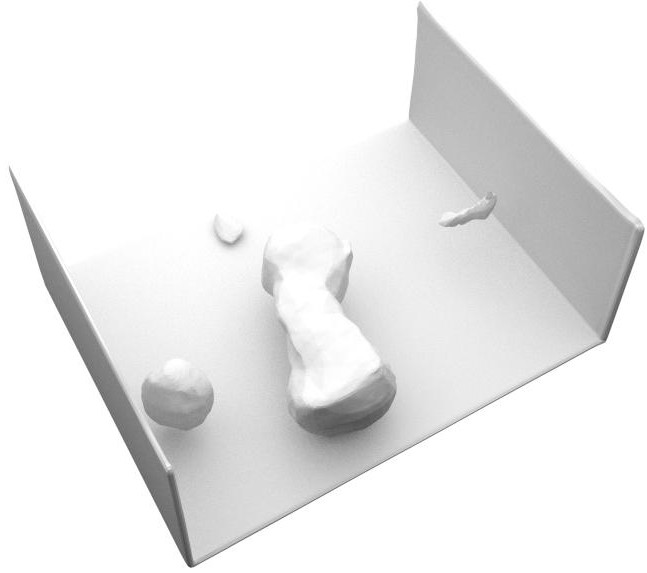} &		
		\includegraphics[width=\mywidths]{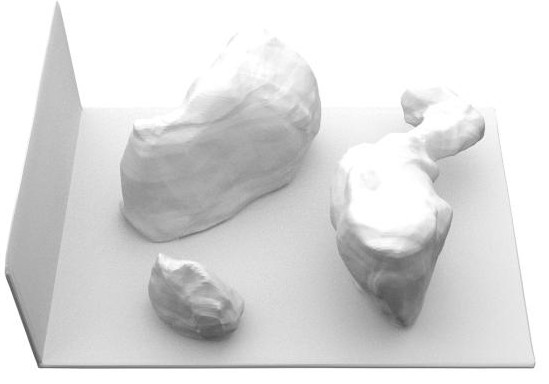}
		\\[-5pt]		
		\rotatebox{90}{PointConv} &&		
		\includegraphics[width=\mywidth]{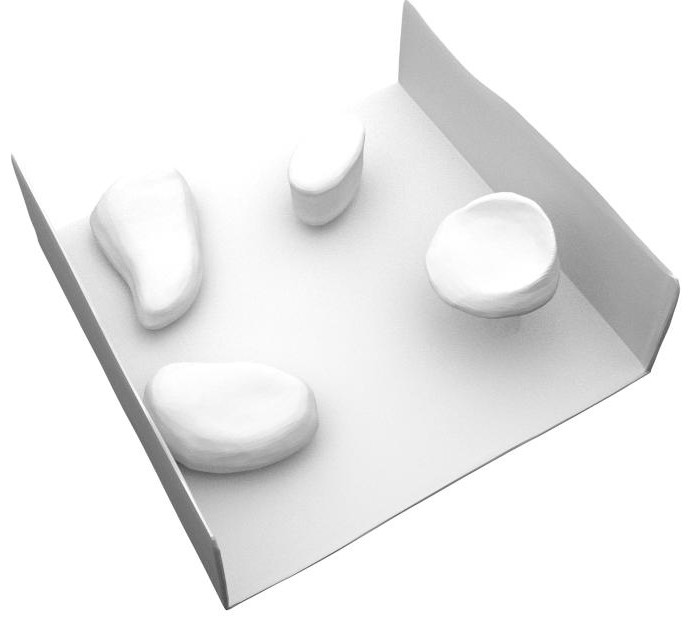} &
		\includegraphics[width=\mywidth]{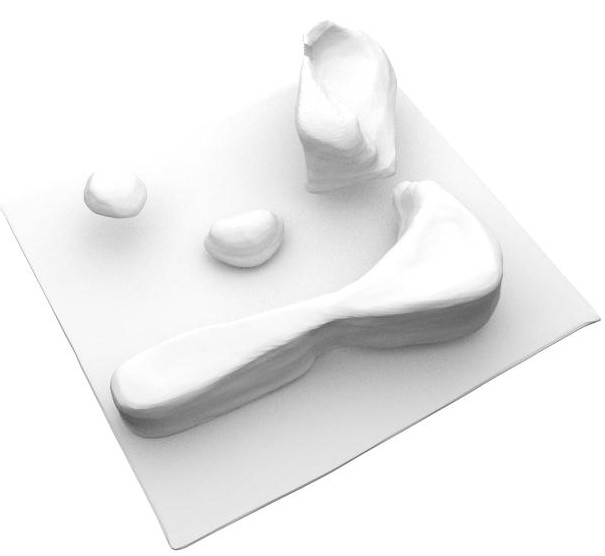} & 
		\includegraphics[width=\mywidth]{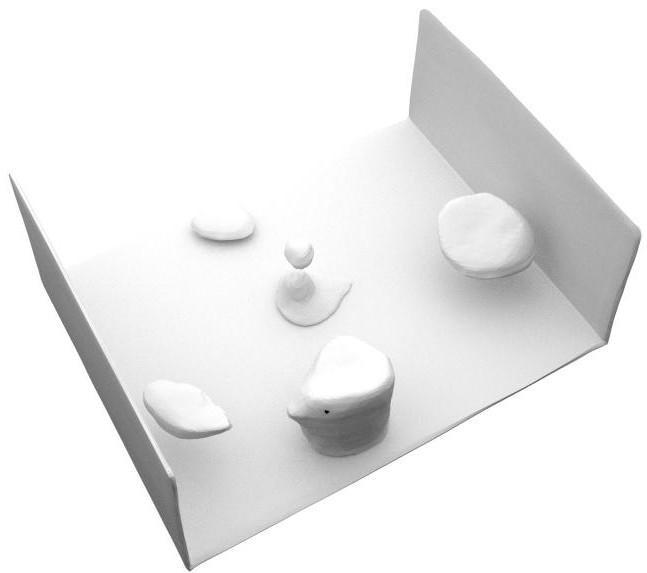} &		
		\includegraphics[width=\mywidth]{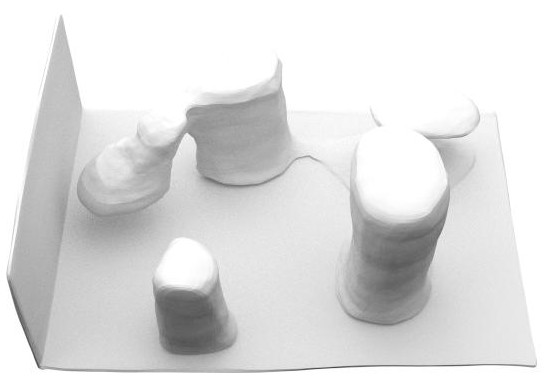}
		\\[-5pt]
		\rotatebox{90}{SPSR~\cite{Kazhdan2013SIGGRAPH}} &\rotatebox{90}{(trimmed)}&
		\includegraphics[width=\mywidth]{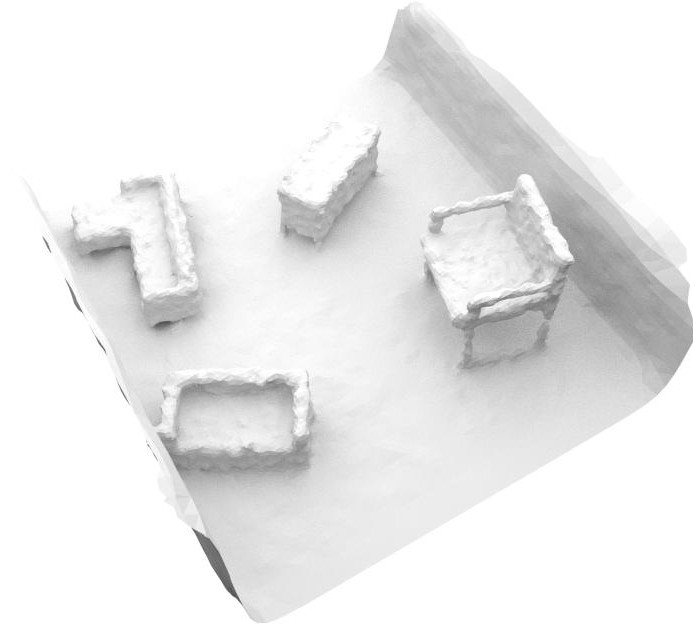} &
		\includegraphics[width=\mywidth]{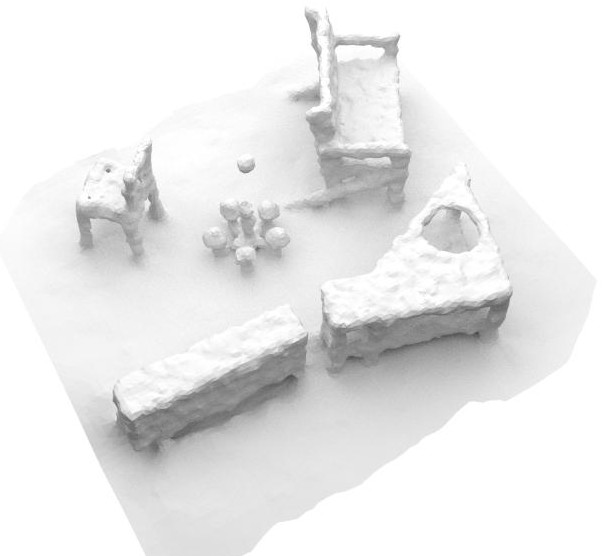} & 
		\includegraphics[width=\mywidth]{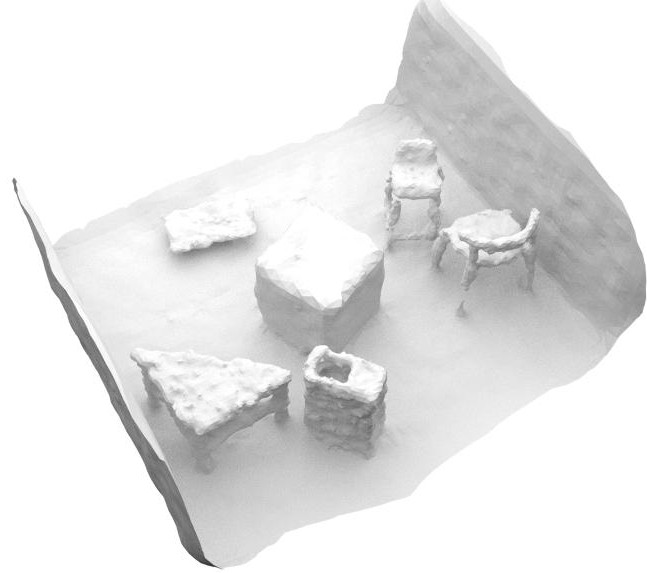} &		
		\includegraphics[width=\mywidth]{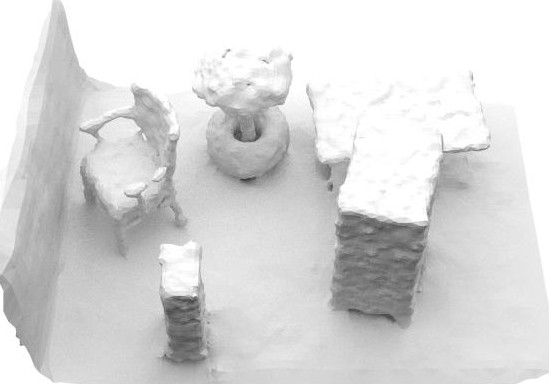}
		\\[-2pt]
		\rotatebox{90}{Ours-2D} & \rotatebox{90}{($3\times128^2$)} &
		\includegraphics[width=\mywidth]{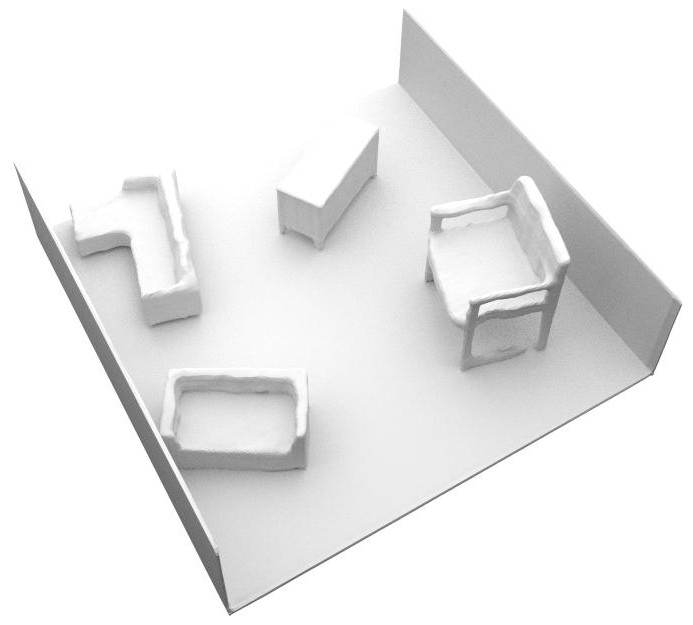} &
		\includegraphics[width=\mywidth]{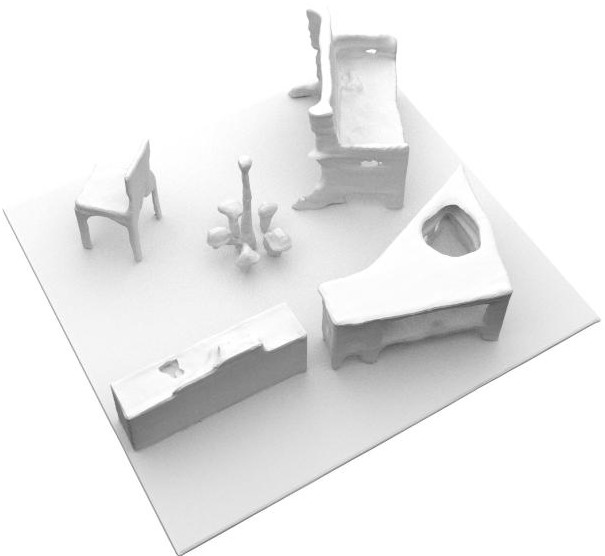} & 
		\includegraphics[width=\mywidth]{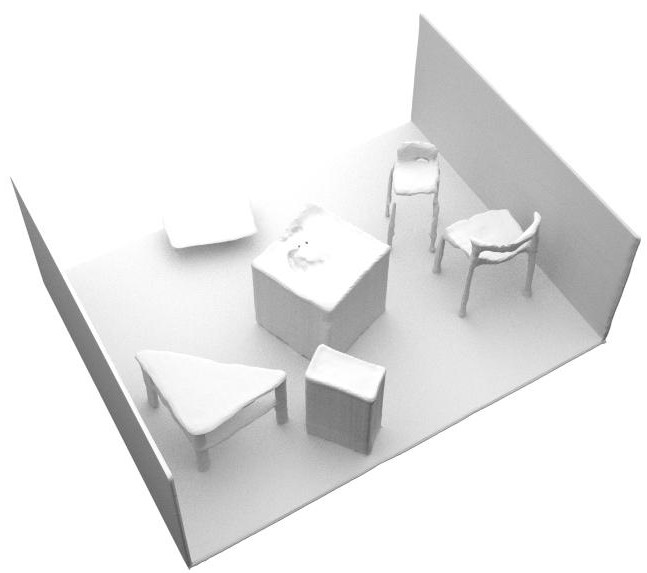} &		
		\includegraphics[width=\mywidth]{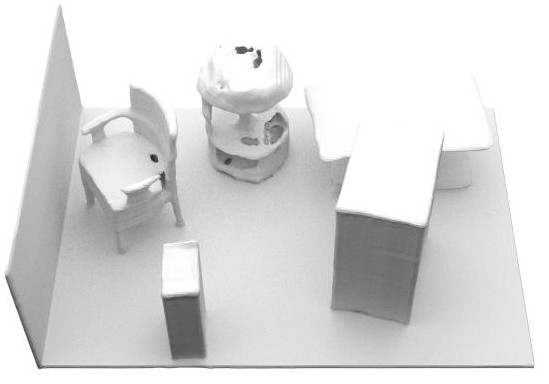}
		\\[-2pt]
		\rotatebox{90}{Ours-3D} & \rotatebox{90}{\quad($32^3$)} &
		\includegraphics[width=\mywidth]{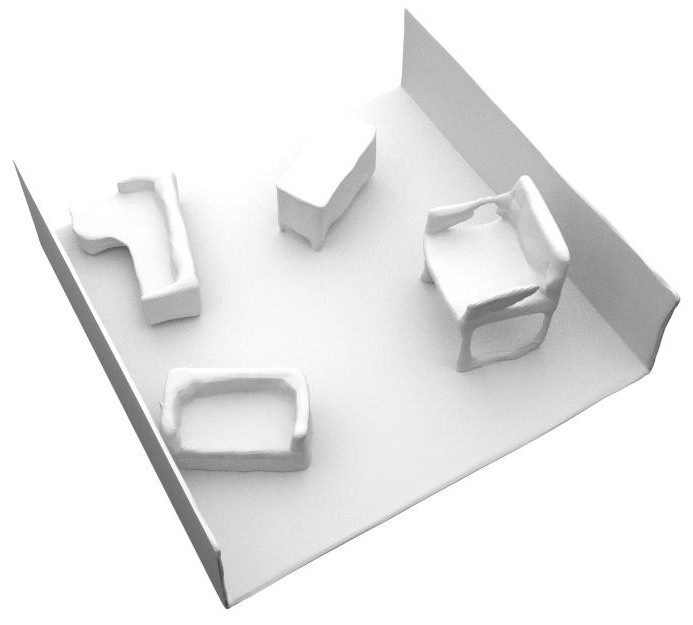} &
		\includegraphics[width=\mywidth]{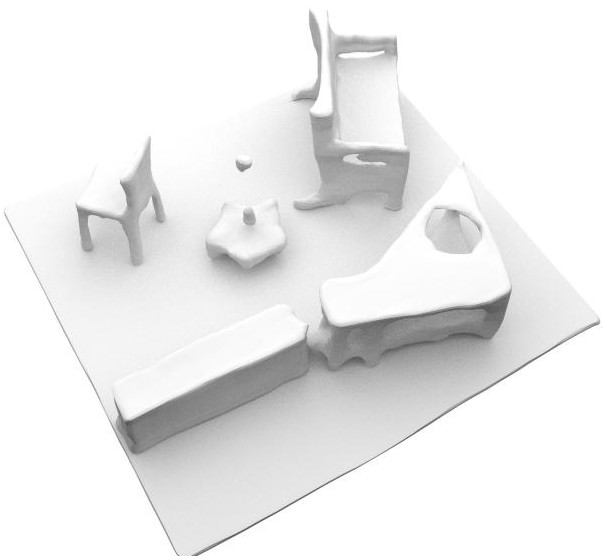} & 
		\includegraphics[width=\mywidth]{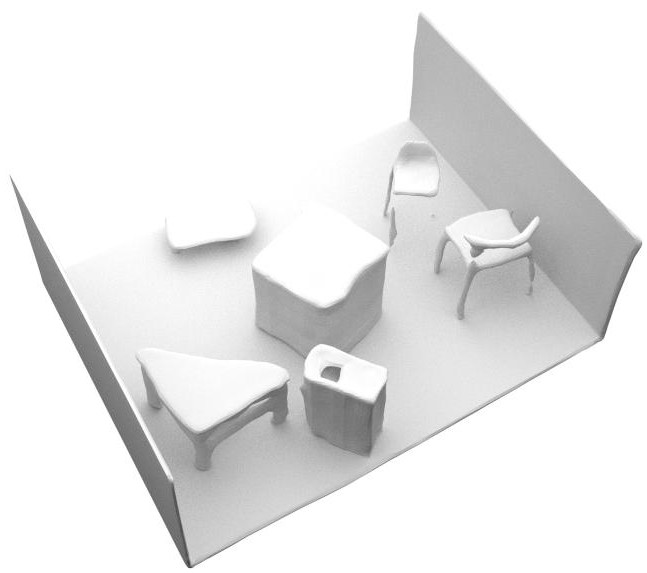} &		
		\includegraphics[width=\mywidth]{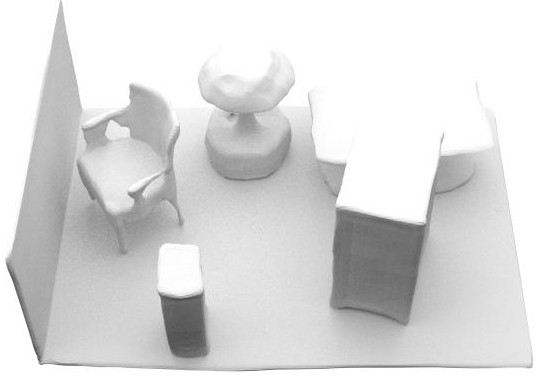}
		\\[-2pt]
		\rotatebox{90}{\quad Ours-2D-3D} & \rotatebox{90}{($3\times128^2 + 32^3$)} &
		\includegraphics[width=\mywidth]{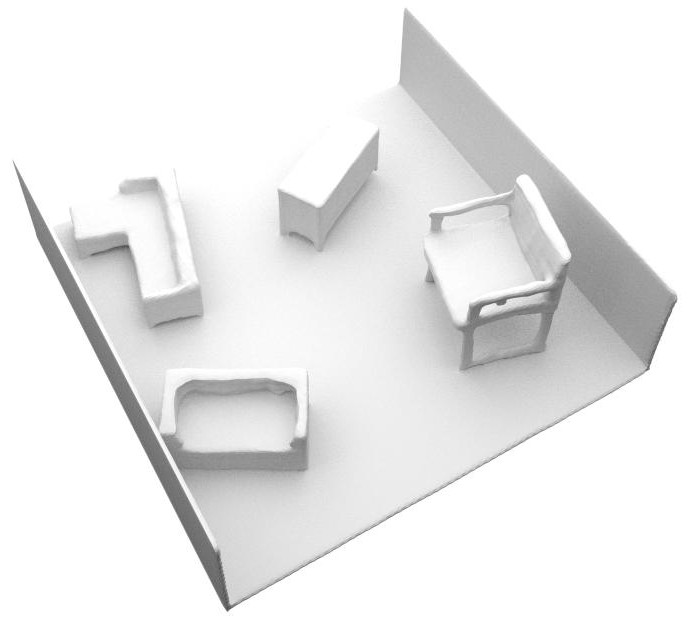} &
		\includegraphics[width=\mywidth]{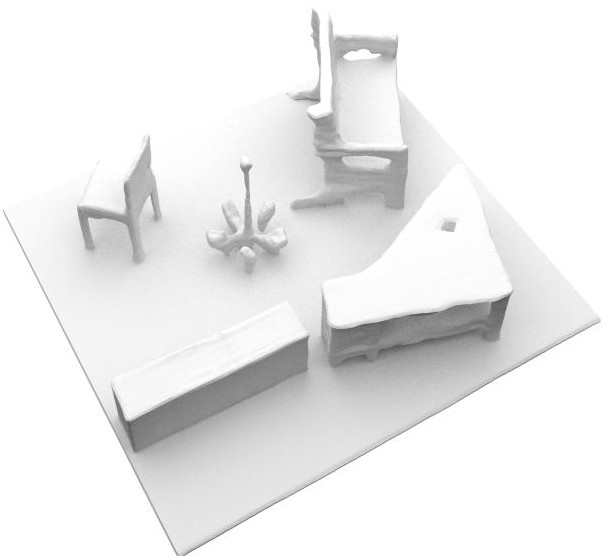} & 
		\includegraphics[width=\mywidth]{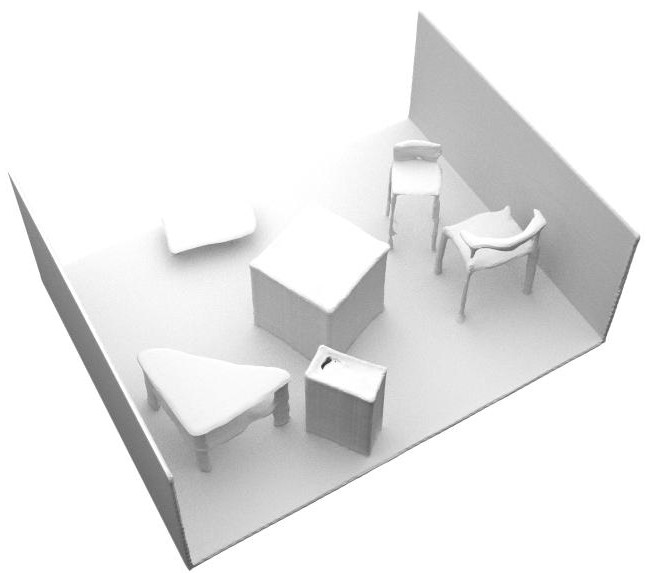} &		
		\includegraphics[width=\mywidth]{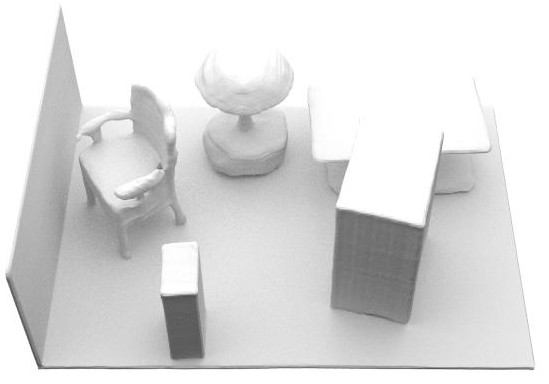}
		\\[-2pt]
		\rotatebox{90}{\quad GT Mesh} &&
		\includegraphics[width=\mywidth]{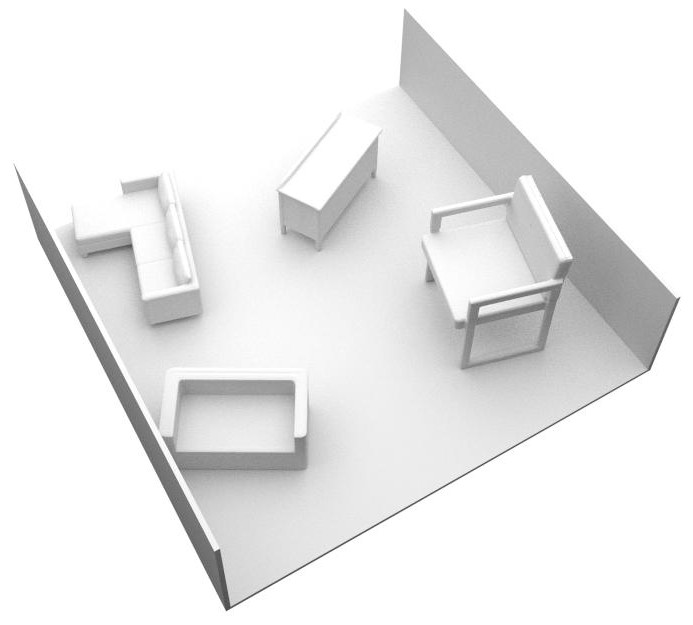} &
		\includegraphics[width=\mywidth]{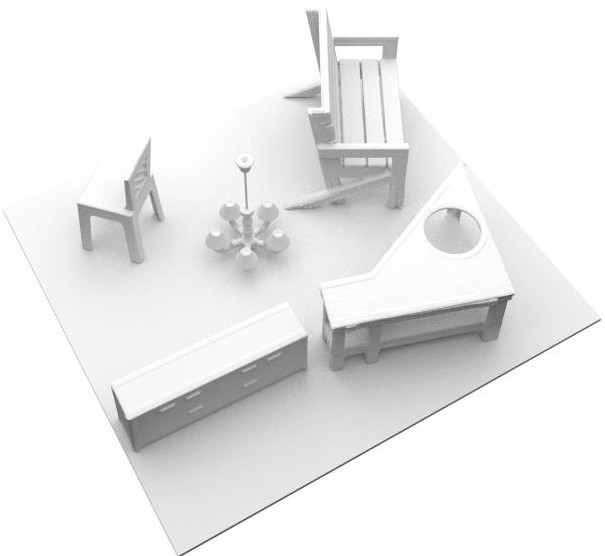} & 
		\includegraphics[width=\mywidth]{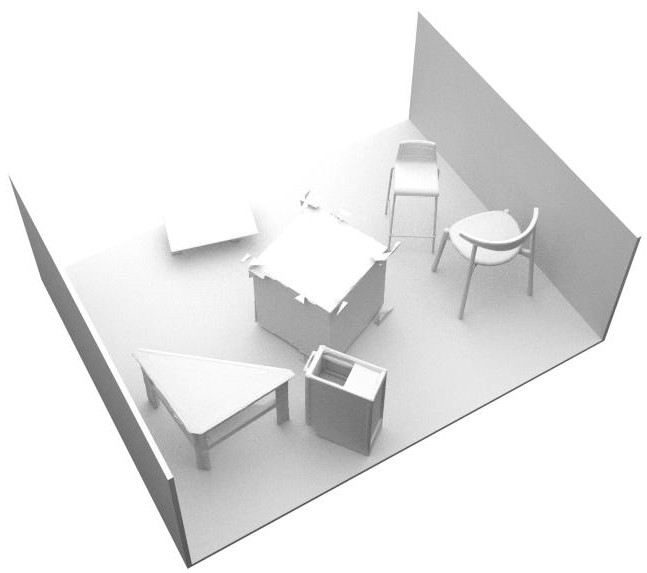} &		
		\includegraphics[width=\mywidth]{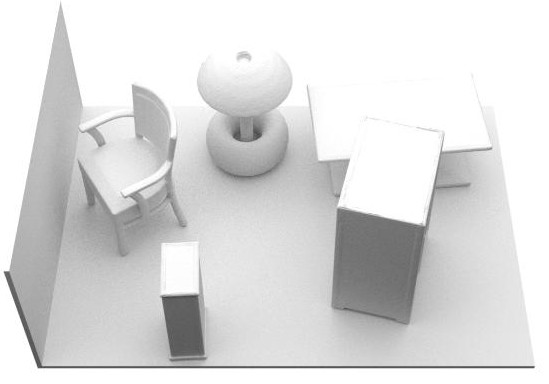}
		\\
	\end{tabular}
	\caption{
		\textbf{Scene-Level Reconstruction on Synthetic Rooms.}
		Qualitative comparison for point-cloud based reconstruction on the synthetic indoor scene dataset. %
	}
	\label{fig:shapenet_synroom}
\end{figure}

\tabref{tab:room} and \figref{fig:shapenet_synroom} show our results.
All variants of our method are able to reconstruct geometric details of the scenes and lead to smooth results.
In contrast, ONet and PointConv suffer from low accuracy while SPSR leads to noisy surfaces.
While high-resolution canonical plane features capture fine details they are prone to noise.
Low-resolution volumetric features are instead more robust to noise, yet produce smoother surfaces. 
Combining complementary volumetric and plane features improves results compared to considering them in isolation.
This confirms our hypothesis that plane-based and volumetric features are complementary.
However, the best results in this setting are achieved when increasing the resolution of the volumetric features to $64^3$.

\begin{table*}[!bt]
	\setlength{\tabcolsep}{0.5cm}
	\centering
	\resizebox{0.9\textwidth}{!}{%
		\begin{tabular}{lcccc}
\toprule
{} &             IoU &   Chamfer-$L_1$ & Normal Consistency & F-Score \\
\midrule
ONet~\cite{Mescheder2019CVPR}      &  0.475 &  0.203 &              0.783 & 0.541\\
PointConv & 0.523& 0.165& 0.811 &0.790\\
SPSR~\cite{Kazhdan2013SIGGRAPH} & - & 0.223& 0.866 & 0.810\\
SPSR~\cite{Kazhdan2013SIGGRAPH} (trimmed) & - & 0.069& 0.890 &0.892\\\hline
Ours-2D ($128^2$)&           0.795 &           0.047 &              0.889 & 0.937\\
Ours-2D ($3\times128^2$)&           0.805 &           0.044 &              0.903 & 0.948	\\
Ours-3D ($32^3$)&          0.782  & 0.047 & 0.902 & 0.941\\
Ours-3D ($64^3$) & \textbf{0.849}& \textbf{0.042}& \textbf{0.915} & \textbf{0.964}\\
Ours-2D-3D ($3\times128^2+32^3$)&           0.816 &           0.044 &              0.905 & 0.952\\
\bottomrule
\end{tabular}

	}
	\caption{
	\textbf{Scene-Level Reconstruction on Synthetic Rooms.}
	Quantitative comparison for reconstruction from noisy point clouds.
	We do not report IoU for SPSR as SPSR generates only a single surface for walls and the ground plane.
	To ensure a fair comparison to SPSR, we compare all methods with only a single surface for walls/ground planes when calculating Chamfer-$L_1$ and F-Score.
}
\label{tab:room}
\vspace{-0.7cm}
\end{table*}

\subsection{Ablation Study}
In this section, we investigate on our synthetic indoor scene dataset different feature aggregation strategies at similar GPU memory consumption as well as different feature interpolation strategies.

\boldparagraph{Performance at Similar GPU Memory}
\tabref{tab:room_ablation_memory} shows a comparison of different feature aggregation strategies at similar GPU memory utilization.
Our multi-plane approach slightly outperforms the single plane and the volumetric approach in this setting. 
Moreover, the increase in plane resolution for the single plane variant does not result in a clear performance boost, demonstrating that
higher resolution does not necessarily guarantee better performance.

\begin{table}[!bt]
	\centering
	\begin{subfigure}{0.57\linewidth}
		\centering
		\resizebox{\textwidth}{!}{%
			\begin{tabular}{lccccc}
\toprule
& GPU Memory     & IoU &   Chamfer-$L_1$ & Normal C. & F-Score\\
\midrule
Ours-2D ($192^2$) &  9.5GB  & 0.773 & 0.047         &   0.889 &0.937\\
Ours-2D ($3\times128^2$) &  9.3GB &         \textbf{0.805} &           \textbf{0.044} &        \textbf{0.903} & \textbf{0.948}\\
Ours-3D ($32^3$) &  8.5GB   & 0.782  & 0.047 & 0.902 &0.941\\
\bottomrule
\end{tabular}

		}\caption{Performance at similar GPU Memory} 
		\label{tab:room_ablation_memory}
	\end{subfigure}
	\begin{subfigure}{0.42\linewidth}
		\centering
		\resizebox{\textwidth}{!}{%
			\begin{tabular}{lcccc}
\toprule
& IoU &   Chamfer-$L_1$ & Normal C. & F-Score \\
\midrule
Nearest Neighbor &  0.766  & 0.052 & 0.885 & 0.920\\
Bilinear &  \textbf{0.805} &           \textbf{0.044} &        \textbf{0.903} & \textbf{0.948}\\
\bottomrule
\end{tabular}

		}
		\vspace{0.27cm}
		\caption{Interpolation Strategy}
		\label{tab:room_ablation_interpolation}
	\end{subfigure}\\
	\vspace{-0.2cm}
	\caption{\textbf{Ablation Study on Synthetic Rooms.} 
	We compare the performance of different feature aggregation strategies at similar GPU memory in~\tabref{tab:room_ablation_memory} and evaluate two different sampling strategies in~\tabref{tab:room_ablation_interpolation}.
	}
\end{table}

\boldparagraph{Feature Interpolation Strategy}
To analyze the effect of the feature interpolation strategy in the convolutional decoder of our method, we compare nearest neighbor and bilinear interpolation for our multi-plane variant.
The results in \tabref{tab:room_ablation_interpolation} clearly demonstrate the benefit of bilinear interpolation.

\subsection{Reconstruction from Point Clouds on Real-World Datasets}
Next, we investigate the generalization capabilities of our method.
Towards this goal, we evaluate our models trained on the synthetic indoor scene dataset on the real world datasets ScanNet~v2~\cite{Dai2017CVPR} and Matterport3D~\cite{Chang2017THREEDV}.
Similar to our previous experiments, we use $10000$ points sampled from the meshes as input.

\boldparagraph{ScanNet v2}
Our results in~\tabref{tab:scannet} show that
among all our variants, the volumetric-based models perform best, %
indicating that the plane-based approaches are more affected by the domain shift.
We find that 3D CNNs are more robust to noise as they aggregate features from all neighbors which results in smooth outputs.
Moreover, all variants outperform the learning-based baselines by a significant margin.

\begin{table*}[!tb]
	\centering
	\resizebox{0.9\textwidth}{!}{%
		\begin{tabular}{lcc}
\toprule
{} &             Chamfer-$L_1$  & F-Score\\
\midrule
ONet~\cite{Mescheder2019CVPR}      &  0.398 & 0.390 \\
PointConv & 0.316 & 0.439\\
SPSR~\cite{Kazhdan2013SIGGRAPH} & 0.293 & 0.731\\
SPSR~\cite{Kazhdan2013SIGGRAPH} (trimmed) & 0.086& 0.847\\
\bottomrule\\
\end{tabular}
\hspace{1cm}
\begin{tabular}{lcc}
	\toprule
	{} &             Chamfer-$L_1$ & F-Score \\
	\midrule
	Ours-2D ($128^2$)&           0.139 & 0.747\\
	Ours-2D ($3\times128^2$)&     0.142 & 0.776\\
	Ours-3D ($32^3$)&          0.095 & 0.837\\
	Ours-3D ($64^3$)&          \textbf{0.077} & \textbf{0.886} \\
	Ours-2D-3D ($3\times128^2+32^3$)& 0.099 & 0.847\\
	\bottomrule
\end{tabular}
	}
	\vspace{0.2cm}
	\caption{
	\textbf{Scene-Level Reconstruction on ScanNet.}
	Evaluation of point-based reconstruction on the real-world ScanNet dataset. As ScanNet does not provide watertight meshes, we trained all methods on the synthetic indoor scene dataset.
	Remark: In ScanNet, walls / floors are only observed from one side. To not wrongly penalize methods for predicting walls and floors with thickness (0.01 in our training set), we chose a F-Score threshold of 1.5\% for this experiment.
}
\label{tab:scannet}
\end{table*}

\begin{figure}[!tb]
	\centering
	\newcommand{\mywidth}{0.32\linewidth}	
	\newcommand{\mywidths}{0.25\linewidth}		
	\newcommand{\mywidthss}{0.29\linewidth}		
	\begin{tabular}{ccccc}
		\rotatebox{90}{\qquad Input}&&
		\includegraphics[width=\mywidth]{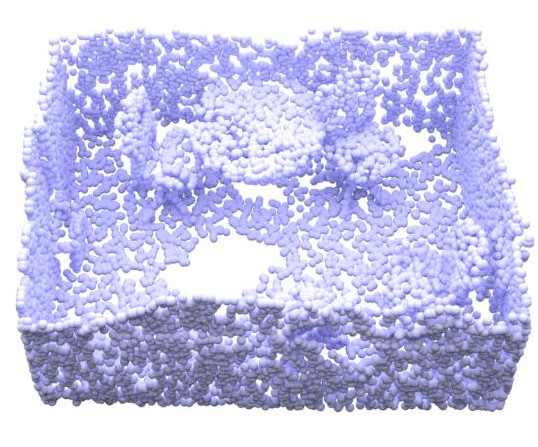} &
		\includegraphics[width=\mywidthss]{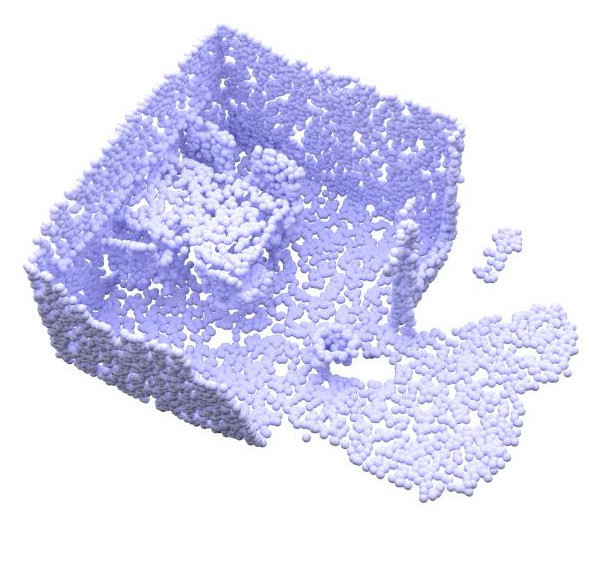}&
		\includegraphics[width=\mywidths]{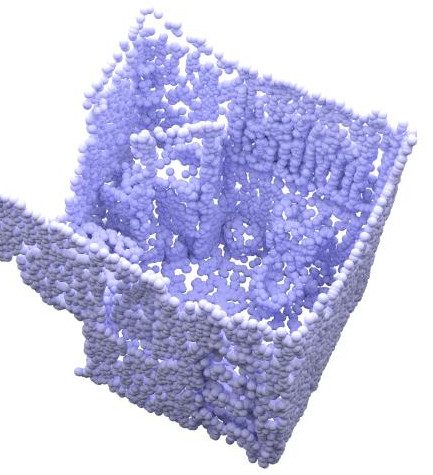}   \\
		\rotatebox{90}{\qquad ONet~\cite{Mescheder2019CVPR}} &&
		\includegraphics[width=\mywidth]{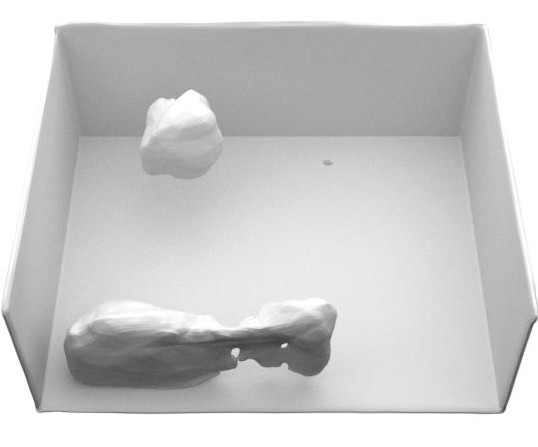} &
		\includegraphics[width=\mywidthss]{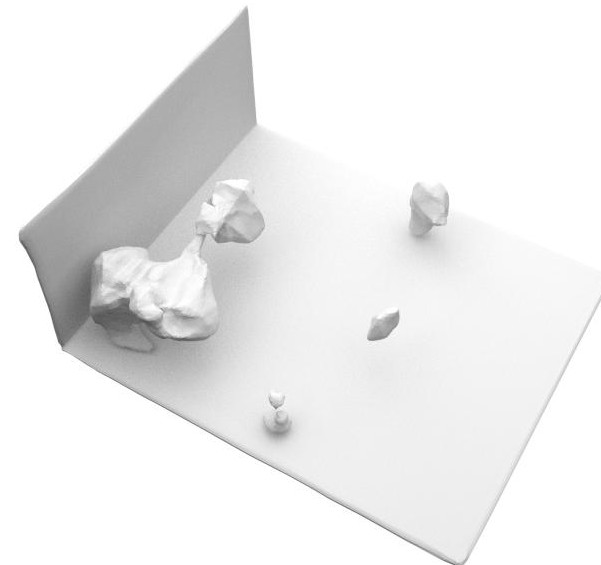} &		
		\includegraphics[width=\mywidths]{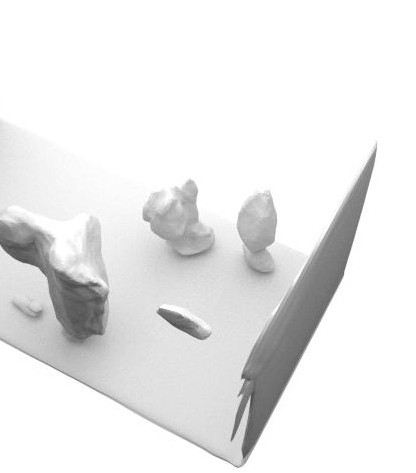} \\
		\rotatebox{90}{\qquad SPSR~\cite{Kazhdan2013SIGGRAPH}} &\rotatebox{90}{\qquad(trimmed)}&
		\includegraphics[width=\mywidth]{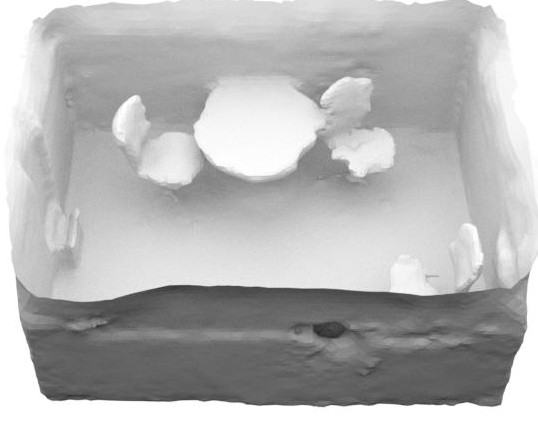} &
		\includegraphics[width=\mywidthss]{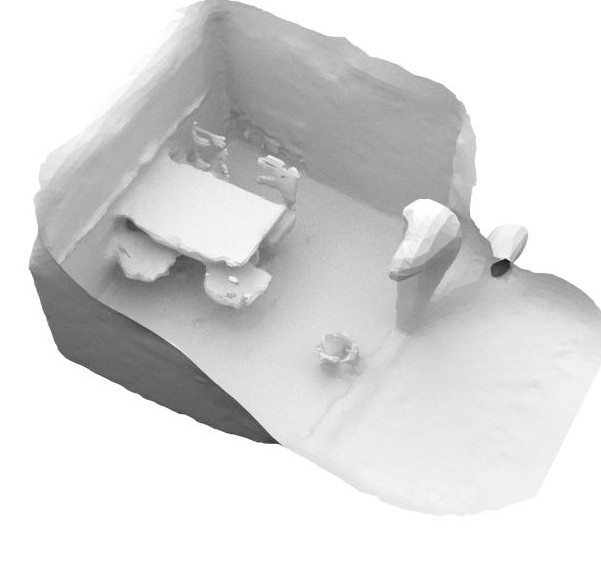} &		
		\includegraphics[width=\mywidths]{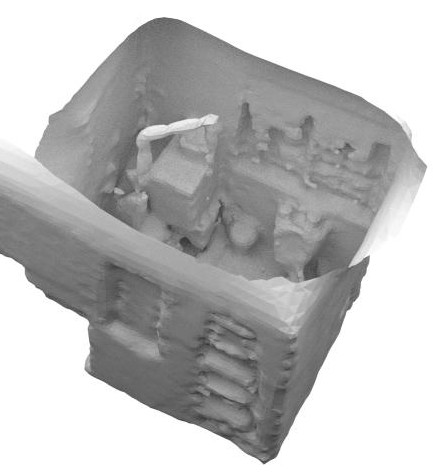} \\
		\rotatebox{90}{\qquad Ours-2D-3D} & \rotatebox{90}{\quad($3\times128^2 + 32^3$)}&		
		\includegraphics[width=\mywidth]{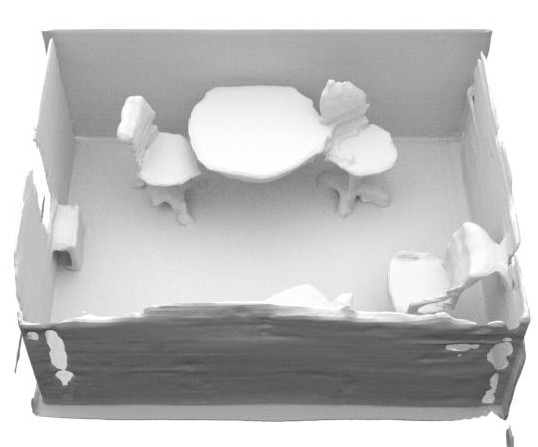} &
		\includegraphics[width=\mywidthss]{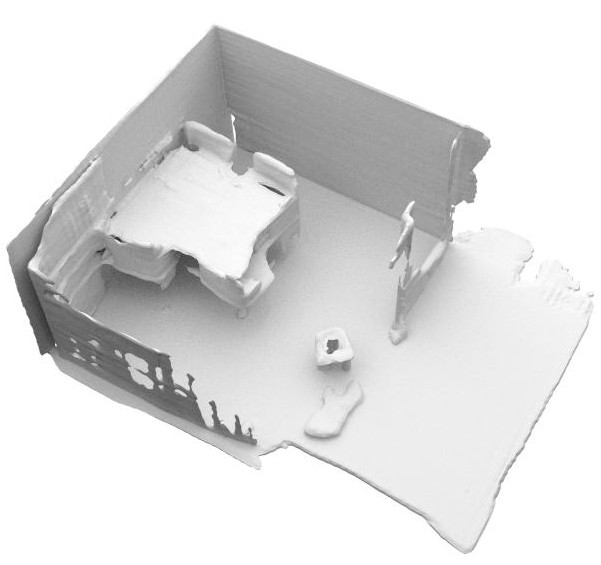}& 		
		\includegraphics[width=\mywidths]{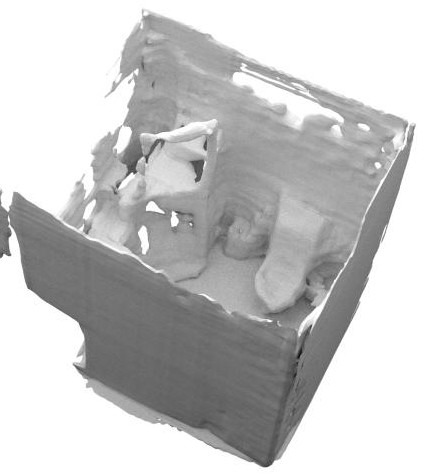}\\
		\rotatebox{90}{\quad Ours-3D} & \rotatebox{90}{\qquad($64^3$)}&		
		\includegraphics[width=\mywidth]{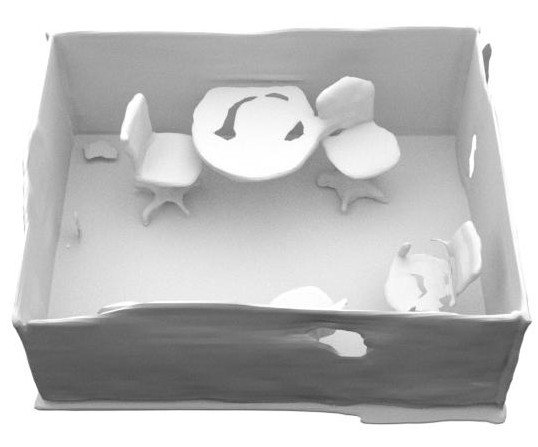} &
		\includegraphics[width=\mywidthss]{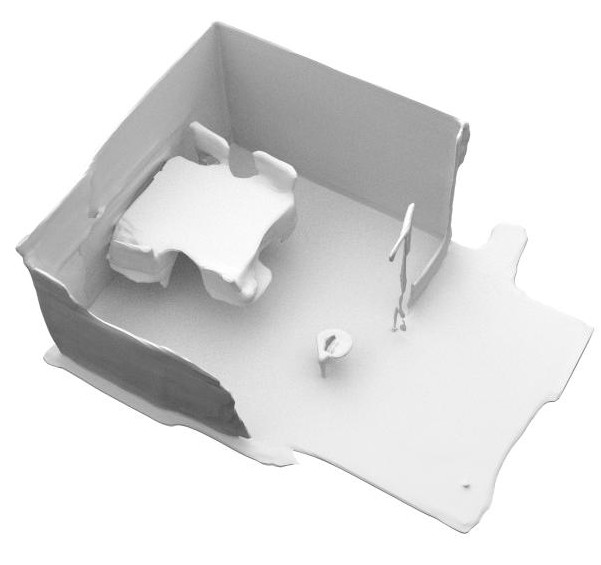}& 				
		\includegraphics[width=\mywidths]{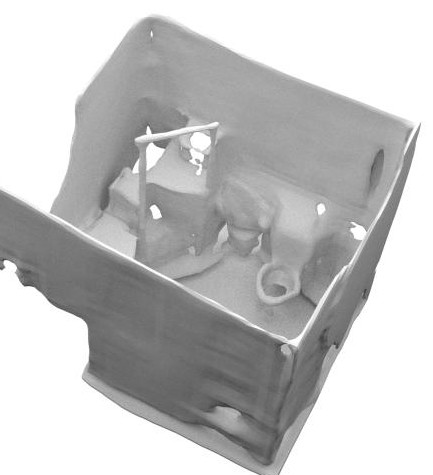} \\
	\end{tabular}
	\caption{
		\textbf{Scene-Level Reconstruction on ScanNet.}
		Qualitative results for point-based reconstruction on ScanNet~\cite{Dai2017CVPR}.
		All learning-based methods are trained on the synthetic room dataset and evaluated on ScanNet.
	}
	\label{fig:scannet_syn2real}
	\vspace{-0.5cm}
\end{figure}

The qualitative comparison in~\figref{fig:scannet_syn2real} shows that our model is able to smoothly reconstruct scenes with geometric details at various scales. %
While Screened PSR~\cite{Kazhdan2013SIGGRAPH} also produces reasonable reconstructions, it tends to close the resulting meshes and hence requires a carefully chosen trimming parameter.
In contrast, our method does not require additional hyperparameters.

\boldparagraph{Matterport3D Dataset}
Finally, we investigate the scalability of our method to larger scenes which comprise multiple rooms and multiple floors.
For this experiment, we exploit the Matterport3D dataset.
Unlike before, we implement a fully convolutional version of our 3D model that can be scaled to any size by running on overlapping crops of the input point cloud in a sliding window fashion.
The overlap is determined by the size of the receptive field to ensure correctness of the results.
\figref{fig:teaser} shows the resulting 3D reconstruction.
Our method reconstructs details inside each room while adhering to the room layout.
Note that the geometry and point distribution of the Matterport3D dataset differs significantly from the synthetic indoor scene dataset which our model is trained on.
This demonstrates that our method is able to generalize not only to unseen classes, but also novel room layouts and sensor characteristics.
More implementation details and results can be found in supplementary.

\section{Conclusion}

We introduced Convolutional Occupancy Networks, a novel shape representation which combines the expressiveness of convolutional neural networks with the advantages of implicit representations.
We analyzed the tradeoffs between 2D and 3D feature representations and found that incorporating convolutional operations facilitates generalization to unseen classes, novel room layouts and large-scale indoor spaces.
We find that our 3-plane model is memory efficient, works well on synthetic scenes and allows for larger feature resolutions. Our volumetric model, in contrast, outperforms other variants on real-world scenarios while consuming more memory.

Finally, we remark that our method is not rotation equivariant and only translation equivariant with respect to translations that are multiples of the defined voxel size. Moreover, there is still a performance gap between synthetic and real data.
While the focus of this work was on learning-based 3D reconstruction, in future work, we plan to apply our novel representation to other domains such as implicit appearance modeling and 4D reconstruction.

\vspace{0.2cm}

\boldparagraph{Acknowledgements}
This work was supported by an NVIDIA research gift.
The authors thank Max Planck ETH Center for Learning Systems (CLS) for
supporting Songyou Peng and the International Max Planck Research School for
Intelligent Systems (IMPRS-IS) for supporting Michael Niemeyer.

\clearpage
\bibliographystyle{splncs04}
\bibliography{bibliography_long,bibliography,bib/bibliography_custom}
\end{document}